\documentclass{article} % For LaTeX2e
\usepackage{collas2026_conference,times}
\usepackage{easyReview}

\usepackage{amsmath,amsfonts,bm}

\def\eqref#1{equation~\ref{#1}}
\def\1{\bm{1}}

\def\mW{{\bm{W}}}

\DeclareMathAlphabet{\mathsfit}{\encodingdefault}{\sfdefault}{m}{sl}
\SetMathAlphabet{\mathsfit}{bold}{\encodingdefault}{\sfdefault}{bx}{n}

\usepackage{hyperref}
\hypersetup{
    colorlinks=true,
    linkcolor=red,
    filecolor=magenta,
    urlcolor=blue,
    citecolor=purple,
    pdftitle={Overleaf Example},
    pdfpagemode=FullScreen,
    }

\usepackage[utf8]{inputenc} % allow utf-8 input
\usepackage[T1]{fontenc}    % use 8-bit T1 fonts
\usepackage{url}            % simple URL typesetting
\usepackage{booktabs}       % professional-quality tables
\usepackage{amsfonts}       % blackboard math symbols
\usepackage{amssymb}
\usepackage{amsthm}
\usepackage{algorithm}
\usepackage{nicefrac}       % compact symbols for 1/2, etc.
\usepackage{microtype}      % microtypography
\usepackage{xcolor}         % colors
\usepackage{graphicx}
\usepackage{subcaption}
\usepackage{mathtools}
\usepackage{cleveref}
\usepackage{siunitx}
\usepackage{svg}
\usepackage{wrapfig}
\usepackage{enumitem}
\usepackage{adjustbox}
\usepackage{enumitem}
\usepackage{commath}
\usepackage{relsize}
\usepackage{tabularx}
\usepackage{bbm}
\usepackage{placeins}
\usepackage[most]{tcolorbox}
\usepackage{tikz}
\usetikzlibrary{backgrounds, fit, calc, positioning, shapes, arrows.meta}

\title{Putting a Face to Forgetting: Continual Learning meets Mechanistic Interpretability}
\author{%
  Sergi Masip \\ KU Leuven \And 
  Gido M.\ van de Ven \\ University of Groningen \And 
  Javier Ferrando \\ Amazon\thanks{This work is not related to the author's position at Amazon.} \And 
  Tinne Tuytelaars \\ KU Leuven \and \\
  \centerline{\tt\small\{sergi.masipcabeza, tinne.tuytelaars\}@kuleuven.be, } \and
  \centerline{\tt\small g.m.van.de.ven@rug.nl, jferrandomonsonis@gmail.com}
}

\newcommand{\rebuttal}[1]{#1}

\collasfinalcopy % Uncomment for camera-ready version, but NOT for submission.

\theoremstyle{plain}
\newtheorem{theorem}{Theorem}[section]
\newtheorem{proposition}[theorem]{Proposition}

\newtheorem{corollary}[theorem]{Corollary}
\theoremstyle{definition}
\newtheorem{definition}[theorem]{Definition}

\theoremstyle{remark}
\newtheorem{remark}[theorem]{Remark}

\begin{document}

\tikzset{fontscale/.style = {font=\relsize{#1}}}

\newcommand{\networkfigure}[9]{
    \begin{tikzpicture}[
        node distance=0.25cm and 0.25cm,
        every node/.style={circle},
        rednode/.style={fill={rgb,255:red,239; green,83; blue,80}},
        bluenode/.style={fill={rgb,255:red,66; green,165; blue,245}},
        greennode/.style={fill={rgb,255:red,102; green,187; blue,106}},
        greynode/.style={fill={rgb,255:red,55; green,55; blue,55}}
      ]
    
      % First layer (3 nodes)
      \node[rednode]   (A1) {};
      \node[bluenode, below=of A1] (A2) {};
      \node[greennode, below=of A2] (A3) {};
    
      % Second layer (2 nodes), vertically centered with respect to the 3 input nodes
      \node[greynode, right=0.5cm of A2, yshift=0.3cm] (B1) {};
      \node[greynode, below=of B1] (B2) {};
    
      % Edges from first to second layer (as thick lines)
        \draw[color=#1, line width=1.5pt] (A1) -- (B1);
        \draw[color=#2, line width=1.5pt] (A1) -- (B2);
        \draw[color=#3, line width=1.5pt] (A2) -- (B1);
        \draw[color=#4, line width=1.5pt] (A2) -- (B2);
        \draw[color=#5, line width=1.5pt] (A3) -- (B1);
        \draw[color=#6, line width=1.5pt] (A3) -- (B2);
    
      % Coordinate system and arrows
        \coordinate (origin) at ($(B1)!0.5!(B2)+(1.2,0)$);
        
        % Axis ends
        \coordinate (xpos) at ($(origin) + (0.8, 0)$);
        \coordinate (xneg) at ($(origin) + (-0.8, 0)$);
        \coordinate (ypos) at ($(origin) + (0, 0.8)$);
        \coordinate (yneg) at ($(origin) + (0, -0.8)$);
        
        \draw[line width=1.2pt, dashed, gray] (origin) -- (xpos);
        \draw[line width=1.2pt, dashed, gray] (origin) -- (xneg);
        \draw[line width=1.2pt, dashed, gray] (origin) -- (ypos);
        \draw[line width=1.2pt, dashed, gray] (origin) -- (yneg);
        
        % Arrow ends
        \coordinate (arrow1) at ($(origin) + #7$);
        \coordinate (arrow2) at ($(origin) + #8$);
        \coordinate (arrow3) at ($(origin) + #9$);
        
        \draw[-{Triangle[width=6pt, length=6pt]}, line width=2pt, color={rgb,255:red,239; green,83; blue,80}]   (origin) -- (arrow1);
        \draw[-{Triangle[width=6pt, length=6pt]}, line width=2pt, color={rgb,255:red,66; green,165; blue,245}]  (origin) -- (arrow2);
        \draw[-{Triangle[width=6pt, length=6pt]}, line width=2pt, color={rgb,255:red,102; green,187; blue,106}] (origin) -- (arrow3);

      % Background layer for rounded box
      \begin{pgfonlayer}{background}
        \node[
          shape=rectangle,
          draw={rgb,255:red,239; green,83; blue,80},
          line width=2pt,
          rounded corners,
          fill=red,
          fill opacity=0.05,
          fit=(A1) (A2) (A3) (B1) (B2) (origin) (xpos) (xneg) (ypos) (yneg) (arrow1) (arrow2) (arrow3),
          inner xsep=4pt, 
          inner ysep=6pt, 
          name=box
        ] {};
      \end{pgfonlayer}
      
      % \node at (current bounding box.north) [yshift=0.5cm, shape=rectangle, text centered] {\textbf{#10}};
    \end{tikzpicture}
}

\maketitle

\begin{abstract}
    Catastrophic forgetting in continual learning is often measured at the performance or last-layer representation level, overlooking the underlying mechanisms. We introduce a mechanistic framework that offers a geometric interpretation of catastrophic forgetting as the result of transformations to the encoding of individual features. These transformations can lead to forgetting by reducing the allocated capacity of features or by disrupting their readout by downstream computations. Analysis of a tractable toy model formalizes this view, allowing us to identify best- and worst-case scenarios. Through experiments on this model, we empirically test our formal analysis and highlight the detrimental effect of depth. Finally, we demonstrate how our framework can be used in the analysis of practical models through the use of crosscoders. We do so through a case study example of a Vision Transformer trained on sequential CIFAR-10. Our work provides a new, feature-centric vocabulary for continual learning. Code available at: https://github.com/Atenrev/mechanistic\_forgetting
\end{abstract}

\section{Introduction}

When a neural network trained on one task learns a second, it often catastrophically forgets the first. Current theoretical works largely study this failure from a macroscopic perspective, by analyzing high-level conflicts like task-vector misalignment \citep{wanunderstanding}, using centered kernel alignment \citep{kornblith2019similarity} as a representation similarity measure \citep{ramasesh2021anatomy}, or studying the role of architectural factors like network width \citep{Goldfarb2025ARXIV_Analysis_of_Overparameterization, guha2024diminishing, mirzadeh2022wide}. These studies characterize when interference occurs and what factors correlate with it, but a description of the internal representational transformations associated with forgetting is still lacking. %We provide a comprehensive discussion of related work in \cref{apx:related_work}.

% Add somehwere "This contrasts with..." to better position

% When a neural network trained on one task learns a second, it often catastrophically forgets the first. Current theoretical understanding largely frames this failure as knowledge overwriting or a high-level conflict between tasks, e.g., analyzing the misalignment of task vectors \citep{wanunderstanding} or representational subspaces \citep{ramasesh2021anatomy}. While valuable, this macroscopic view obscures the rich, internal dynamics of forgetting. It explains that tasks interfere, but not precisely how that interference dismantles the network's internal building blocks.
% Mention works on architecture, width, and link with capacity. Also mention those works that study representations through SVD, as they can't capture all the features encoded in the activation if there is superposition.

To fill this gap, our primary contribution is a conceptual framework that offers a granular, mechanistic account of forgetting. We argue that to better understand forgetting, it is valuable to adopt tools that look inside the black box. The field of mechanistic interpretability provides such a lens by aiming to reverse-engineer the algorithms learned by neural networks~\citep{ferrando2024primerinnerworkingstransformerbased,bereska2024mechanistic}. Its core premise is that knowledge is encoded in a network by hierarchically representing many individual \emph{features}---the fundamental units of representation---which correspond to meaningful concepts or patterns in the data.

Under the mechanistic lens, features are typically assumed to be encoded as \emph{linear directions in activation space}~\citep{olah2020zoom, elhage2022toy}. For each feature, a \emph{feature vector} indicates the direction and strength with which that feature is encoded. %In contrast to what is sometimes assumed in continual learning (e.g., EWC~\citep{kirkpatrick2017overcoming}, MAS~\citep{aljundi2018memory}), 
Features rarely map cleanly to individual neurons. Neurons are often polysemantic (i.e.,~encoding multiple features), and networks can represent more features than neurons via \emph{superposition}~\citep{elhage2022toy}. Superposition translates into a crowded, entangled representation where features compete for {\em allocated capacity}~\citep{elhage2022toy, scherlis2022polysemanticity}, a geometric measure of how `cleanly' they are represented. A feature with lower allocated capacity has greater overlap with other features, resulting in a noisier readout (see \cref{sec:rep_read}).

% Features rarely map cleanly to individual neurons. Empirical evidence shows that neurons are often \emph{polysemantic}, encoding multiple, seemingly unrelated concepts at once. Furthermore, networks can represent far more features than they have neurons, a phenomenon known as \emph{superposition} \citep{elhage2022toy}. This forces features into an entangled, overcomplete representation. Consequently, the most robust way to understand a feature is not as a single neuron's activity, but as a \emph{direction in the high-dimensional activation space}~\citep{olah2020zoom, elhage2022toy}.

% This geometric view has a profound implication: in a crowded representational space, features must compete for resources. The "cleanliness" of a feature's representation can be quantified by its \emph{allocated capacity}~\citep{elhage2022toy, scherlis2022polysemanticity}, a measure determined by the geometry of its vector relative to others. A feature vector that is orthogonal to all others has high capacity. In contrast, one that overlaps significantly with others has low capacity and is more difficult for the network to utilize.

In this work, we model forgetting as the effect of transformations applied to feature vectors---in particular, rotations and scaling. As we explain in \cref{sec:effect_capacity}, rotations can increase \emph{overlap} with other features, while scalings can cause \emph{fading}, both of which reduce a feature's allocated capacity, compromising its readability. 
%This means that downstream computations that depend on the feature can no longer extract its information effectively. 
In addition, even if the allocated capacity is not reduced, these transformations can cause {\em readout misalignment}, where downstream computations no longer extract the information about the feature correctly. Our proposed framework provides a new vocabulary to describe the dynamics underlying forgetting in neural networks.

To illustrate the value of our framework, in \cref{sec:dynamics}, we use a tractable toy model to identify best- and worst-case scenarios for feature forgetting during continual learning. We then validate our analytical predictions with an implementation of the tractable model in \cref{sec:toy_experiments}. These results provide a fresh explanation for previously observed phenomena related to task similarity~\citep{lee2021continual, hiratani2024disentangling, doan2021theoretical} and architectural choices~\citep{guha2024diminishing, lu2024revisiting, mirzadeh2022wide}. Finally, in \cref{sec:scalingup}, we demonstrate how our framework can be used in practice. Using \emph{crosscoders}~\citep{lindsey2025crosscoders}, we trace the evolution of features within a Vision Transformer on a standard continual learning vision benchmark, opening new avenues for mechanistic research in continual learning. 

We provide a comprehensive discussion of related work in \cref{apx:related_work}.
\section{Background and terminology}

\begin{figure*}[t]
  \centering
  \includegraphics[width=1\textwidth]{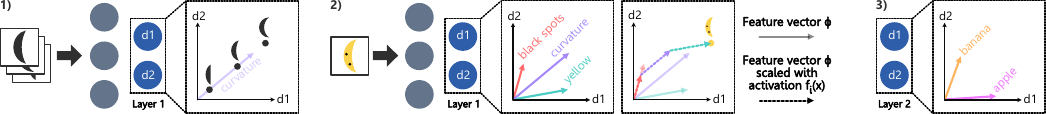}
  \caption{\textbf{Linear representation hypothesis.} 
  % (1) Features are represented as directions ($\phi_i$) in the activation space of a layer ($a_\ell(x)$). (2) The activation of a feature $f_i(x)$ given an input $x$ is encoded linearly along its direction $\phi_i$. (3) The layer’s activation for input $x$ is expressed as a linear combination of feature vectors $\phi_i$, each weighted by its feature activation $f_i(x)$.
  (1)~A feature is a specific concept or pattern in the data (e.g.,~curvature) that is encoded by a linear direction in a layer's activation space. The direction and strength with which a feature is represented are captured by a feature vector. (2)~These feature vectors form the basis for the layer’s representation: its activation in response to an input~$\bm x$ is a linear combination of feature vectors~$\bm{\phi}_i$ weighted by their activations~$f_i(\bm{x})$. (3)~Deeper layers encode increasingly abstract features.
  }
  \label{fig:lrh}
\end{figure*}
Here, we outline the mechanistic foundations underlying our conceptual framework. First, we explain how the term `feature' is used in this paper~(\cref{sec:features}), after which we examine how such features occupy representational capacity and how their encodings are read by downstream computations~(\cref{sec:rep_read}).

\subsection{Features, the building blocks of representation}\label{sec:features}
% Tinne: maybe discuss the effect of different non-linearities.
A neural network transforms input data into rich numerical representations. With increasing depth, these representations capture increasingly abstract structures and concepts or patterns. In the mechanistic literature, these concepts encoded by the different layers of a neural network are typically called \emph{features}. This means that a feature corresponds to a specific, often highly non-linear, pattern in the input data that is encoded by a particular layer of a neural network.
% For a neural network trained to recognize different fruits, features of an intermediate layer might be “curvature”, “yellow”, and “black spots”, while “banana” might be a feature of the penultimate layer.

Our working assumption is the \emph{linear representation hypothesis}~\citep{mikolov2013linguistic, elhage2022toy, bereska2024mechanistic} (\cref{fig:lrh}). It posits that features are encoded as linear directions in a layer's activation space, and that a layer’s activation is a linear combination of the encoded features of that layer: % and ii) each layer's activation can be expressed as a linear combination of the activations of all the features that the model has learned to represent in that layer:
\begin{equation}\label{eq:activation_linear}
\bm{a}_\ell(\bm{x}) \;=\; \Phi_\ell \bm{f}_\ell(\bm x) \;=\;\sum_{i=1}^{n_\ell} f_{\ell,i}(\bm{x}) \, \bm\phi_{\ell,i}.
\end{equation}
Here, $\bm{a}_\ell(\bm{x}) \in \mathbb{R}^{m_\ell}$ represents the activation vector for $m_\ell$ neurons in layer~$\ell$ in response to input~$\bm{x}$. Given $n_\ell$ features encoded by this layer, the \emph{feature vector} $\bm\phi_{\ell,i} \in \mathbb{R}^{m_\ell}$ defines the direction and strength with which the $i^{\text{th}}$ feature of layer~$\ell$ is encoded in the activation, and $f_{\ell,i}(\bm{x})$ denotes the \emph{feature activation}, a scalar coefficient measuring how strongly or to what extent the pattern of the $i^{\text{th}}$ feature of layer~$\ell$ is present in input~$\bm{x}$. When it is clear from the context, we drop the subscript~$\ell$. %In \cref{sec:scalingup}, we show how we can identify features encoded by deep neural network models that are used in practice.

\subsection{Representing and reading features}\label{sec:rep_read}

%Since downstream layers or classifiers rely on these features to make predictions, we n
Next, we examine how features are represented and read out. Two aspects matter:
\begin{enumerate}[topsep=0pt,itemsep=0ex,partopsep=0ex,parsep=1ex, wide, labelwidth=!, labelindent=0pt]
\item \textbf{Allocated capacity:} How much of a layer’s representational space is allocated to each feature?
\item \textbf{Readout quality:} How effectively do downstream computations extract information about each encoded feature?
\end{enumerate}
\noindent
Each layer in a neural network has a finite \emph{representational capacity}; it is not possible to linearly represent an unlimited number of features simultaneously without interference. The geometry of the set $\{\bm{\phi}_{\ell,i}\}_{i=1}^{n_\ell}$---in particular the norms (\emph{strength}) and pairwise cosine similarities (\emph{overlap})---determines the representational capacity allocated to each feature~\citep{elhage2022toy}. 

Features compete for resources, and when they overlap (i.e., ~$\bm \phi_i^\top \bm \phi_j \ne 0$), they share capacity. In practice, features often overlap due to \emph{superposition}~\citep{elhage2022toy}, a phenomenon that arises when the number of represented features exceeds the available representation dimensions. %Overlap can also arise from continual learning, as discussed in \cref{sec:effect_capacity}.
% Candidate to be trimmed?
% For instance, if the "yellow" feature has its own orthogonal dimension, it is represented cleanly. However, if its direction overlaps with that of "curvature" (e.g., there are not enough dimensions), the two features share capacity, reducing how clearly each is encoded. The one with the larger norm has a higher capacity. 

The representational capacity given to a feature determines its \emph{inherent readability}---how well it can be recovered from the activation by a linear readout. We quantify this using the \emph{allocated capacity} $C_i$~\citep{elhage2022toy,scherlis2022polysemanticity}. It captures the "exclusivity" of a feature's encoding:
\begin{equation}\label{eq:allocated_capacity}
C_i = 
\begin{cases}
\dfrac{(\bm\phi_i^\top \bm\phi_i)^2}{\sum_j (\bm\phi_i^\top \bm\phi_j)^2} & \text{if } \|\bm\phi_i\|>0, \text{ else } 0.
\end{cases}
\end{equation}
When $\bm{\phi}_i$ is orthogonal to all other feature vectors $\bm\phi_j$ in a layer, the feature has exclusive access to its dimension. When feature vectors overlap, their $C_i$ decreases, reflecting shared capacity.
% When $\bm{\phi}_i$ is orthogonal to all other feature vectors $\phi_j$, the allocated capacity of that feature is one (it has exclusive access to its dimension). The more the vectors overlap, the lower the capacity, reflecting the sharing of capacity.

Suppose a "yellow" feature~$\bm\phi_i$ is represented in some layer of the network. Downstream computations from subsequent layers must somehow extract this "yellow" information from the layer's representation. We can model this extraction process using a {\em readout vector} $\bm{r}_i$~\citep{hanni2024mathematical} that, for any input~$\bm{x}$, attempts to recover the feature's activation~$f_i(\bm{x})$ from the layer's activation~$\bm{a}(\bm x)$:
\begin{equation}\label{eq:feature_readout}
\bm r_{i}^\top \bm{a}(\bm x) \;\approx\; f_{i}(\bm x).
\end{equation}
The readout vector $\bm{r}_{i}$ controls how well downstream computations read out the feature, but the quality of this readout is fundamentally limited by the inherent readability of a feature's encoding: a perfect linear readout is only possible when its feature vector is orthogonal to all other feature vectors, i.e., $\bm\phi_i^\top \bm\phi_j = 0$ for all $i \neq j$, such that its allocated capacity is one ($C_i=1$). When 
$C_i < 1$,
%a feature has allocated capacity lower than one, 
the feature's readout will be corrupted by noise from overlapping features. 
%In that case, the feature vector itself can serve as the optimal readout vector. 
% Thus, a lower allocated capacity means that a feature becomes harder to read out.  

The distinction between a feature's capacity and its readout may seem unnecessary for standard learning setups, but it is critical for continual learning, as it is key to understanding forgetting in mechanistic terms. It is important to separate a feature's \textbf{inherent readability} (its allocated capacity) from its \textbf{actual readout quality}, as they can become misaligned if downstream computations fail to adapt to changes in the feature's representation during continual learning.
\section{Mechanistic conceptualization of forgetting}\label{sec:conceptualization}

% Having described how features are represented and read out at a single point in time, 
We now turn to how the encoding of features changes as the model continues learning. % Forgetting in continual learning is typically measured as a drop in past-task performance. However, to uncover its mechanism, we examine how the geometry of features evolves.
In this section, we introduce a conceptual framework in which forgetting arises from two primitive transformations to individual feature vectors: \emph{rotation} and \emph{scaling}. From this perspective, we show that forgetting arises from capacity degradation due to feature overlap or fading, and from readout misalignment (\cref{sec:transformations_effects}). With a formal tractable model, we show that these effects jointly define the feature dynamics and enable identifying best- and worst-case scenarios (\cref{sec:dynamics}). 
% We formalize this reasoning with a tractable feature–reader model that permits analytical study of feature evolution and its effect on performance.

\subsection{Transformations of feature vectors}\label{sec:transformations_effects}

\paragraph{Primitive transformations} During continual learning, model parameters are updated, causing feature vectors~$\bm{\phi}_i$ to undergo local transformations that can be decomposed into two primitives: (i) \textbf{rotation} ($\bm \phi_i \to R^\top \bm \phi_i$), which changes the feature vector's direction in representation space, and (ii) \textbf{scaling} ($ \bm \phi_i \to \alpha \bm \phi_i$), which alters its magnitude.

% \paragraph{Primitive transformations}
% During continual learning, feature vectors $\phi_i$ undergo transformations that we can decompose into two primitives:

% \begin{itemize}
%     \item \textbf{Rotation:} $\phi_i \to R^\top \phi_i$ where $R$ is a rotation matrix. The feature changes direction in representation space.
%     \item \textbf{Scaling:} $\phi_i \to \alpha_i \phi_i$ where $\alpha_i \in \mathbb{R}$. The feature changes magnitude.
% \end{itemize}

\paragraph{Effects on allocated capacity}\label{sec:effect_capacity}

A potential consequence of transformations is a change in the network’s allocation of capacity to features. \emph{A degradation in a feature's allocated capacity reduces its inherent readability}. This capacity degradation cannot be recovered by later layers. The allocated capacity can decrease due to:

\begin{itemize}[topsep=0pt,itemsep=0ex,partopsep=0ex,parsep=1ex, wide, labelwidth=!, labelindent=0pt]
    \item \textbf{Overlap:} If $\bm \phi_i$ rotates towards another feature vector~$\bm \phi_j$, then $\bm \phi_i^\top \bm \phi_j$ increases, lowering both $C_i$ and $C_j$.
    \item \textbf{Fading}: Scaling $\bm \phi_i$ changes the strength of a feature’s representation. Scaling affects capacity only when features overlap: reducing its norm (fading) decreases allocated capacity, while increasing its norm (strengthening) boosts its allocated capacity at the expense of others. A feature can thus also lose capacity through strengthening of other overlapping features. In the limit $\|\bm \phi_i\| \to 0$, we say that the feature vanishes, meaning that it is no longer encoded in that layer and can no longer be read out. 
\end{itemize}

\paragraph{Effects on readout}\label{sec:effects_readout}

Even when a feature retains its allocated capacity, \emph{transformations can cause feature vectors and readouts to become misaligned}, impairing the quality of the readout. If $\bm \phi_i$ is scaled or rotated without updating the readout $\bm r_i$, the next layer reads out a weaker, stronger or different signal. However, if the features maintain full allocated capacity, \textit{the resulting performance loss is recoverable} by realigning the readouts. % However, in capacity-constrained settings where features overlap, performance also becomes highly sensitive to the geometric arrangement of features. Each task optimally arranges features for its specific objective (e.g., packing correlated features together as observed in~\citep{elhage2022toy}). Consequently, conflicting updates from new tasks can alter this arrangement, pulling previously grouped features apart and causing an \textit{irrecoverable performance loss} (e.g., by forcing an arrangement where the readout now inevitably reads out noisy interference from other task features).
% However, in capacity-constrained settings with overlapping features, changes in the geometric arrangement can couple features in a way that prevents clean readout separation, limiting the recoverability of this misalignment.

\subsection{Dynamics of features under a feature-reader model}\label{sec:dynamics}

We now investigate how these transformations and their consequences may arise in a continual learning setup under a toy model we term \textit{feature-reader model}. We show how in this model feature preservation depends on task structure, optimizer choice, and representational geometry. Specifically, features absent from a new task receive no gradient pressure and remain stable, while features shared across tasks are preserved when used consistently. In contrast, features that are active but have different contributions across tasks are subject to destructive updates, with the severity amplified by feature co-activation overlap, contribution misalignment, and the number of simultaneous readouts. Weight decay (and possibly other regularized optimizers) and depth further risk degrading the capacity of inactive features.

\paragraph{Model definition}
% In a standard network, a layer performs several computations, generating multiple outputs from the given inputs, i.e., the activation $\bm{a}(\bm x)$ from the previous layer. Specific to our feature-reader model (and similar to the toy autoencoder from \citet{elhage2022toy}) is that we start from the actual feature activations $\bm{f}(\bm x)$, simplifying the analysis. Our model encodes these features using feature vectors $\Phi=[\bm \phi_1,...,\bm \phi_n] \in \mathbb{R}^{m \times n}$.

We introduce a simplified feature-reader model, which serves as an illustration of \textit{how features at a given layer interact with subsequent layers}. Following the same modeling choice as the toy autoencoder model of \citet{elhage2022toy}, we treat the model's input as directly corresponding to a disentangled set of ground-truth feature activations $\bm{f}(\bm x)$. In other words, we assume the identity of the features is known and given by construction, rather than something the model must discover. The model encodes these features linearly in a lower-dimensional embedding via the feature vectors $\Phi=[\bm \phi_1,...,\bm \phi_n] \in \mathbb{R}^{m \times n}$. Unlike \citet{elhage2022toy}'s autoencoder, our model does not attempt to reconstruct the input. Instead, it reads out one or more of the encoded features and linearly combines the readouts. The joint process of reading out the encoded features and linearly combining them is modeled with a \textit{probe} $\bm w \in \mathbb{R}^m$. We start our analysis with a single probe and later extend it to the case of multiple probes. The output of the model is a predicted output label $\hat y(\bm x)$:
% In a standard network, a layer performs several computations, and for each computation, it reads one or more features from the preceding activation. We model all the readouts for one downstream computation along with the way they are combined through a \textit{probe} $\bm{w}\in\mathbb{R}^{m}$. Similar to the toy autoencoder from \citet{elhage2022toy}, our model encodes features using feature vectors $\Phi \in \mathbb{R}^{m \times n}=[\bm \phi_1,...,\bm \phi_n]$, but they are read by a {\em probe} $\bm w \in \mathbb{R}^m$, mapping to a predicted output label $\hat y(\bm x)$:
\begin{equation}\label{eq:model}
    \hat y(\bm x)= \; {\bm w}^\top \Phi \bm{f}(\bm x).
\end{equation}
Each task $T$ assigns a ground-truth label $y^{(T)}(\bm x)$ to each input $\bm x$, which the model is trained to predict. 
% In practice, a task would also be defined by a downstream computation.

For tractability, we assume:
(i)~Feature activations $\bm f(\bm{x})$ are non-negative\footnote{This is consistent with sparse autoencoder implementations where activations are ReLU-constrained, causing some opposing concepts (e.g.,\ dark vs.\ bright) to split into distinct features.}.
(ii)~Feature vectors $\Phi$ are optimized directly, abstracting network parameters.
(iii)~All relevant features for all tasks are known in advance. Thus, the $n$ considered features cover all features learned across all tasks. These assumptions isolate the geometry of feature updates by abstracting away confounding network dynamics. They permit the closed-form derivation of the coupled gradient dynamics in \cref{theorem:updates} and its subsequent corollaries. While relaxing them complicates the formal analysis, we hope that the qualitative conclusions hold, which we back up with empirical examples in \cref{sec:results_toy_depth}.

Note that our model reinterprets the linear model of \citet{saxe2014exact} through the mechanistic lens of superposition \citep{elhage2022toy}. Their SVD-based analysis cannot represent superposition; our formulation explicitly does.

\paragraph{Update mechanisms under continual learning}

With the model defined, we proceed to analyze its update mechanisms. We focus the analysis on 2 tasks without loss of generality. We assume that each task $T\in\{A, B\}$ is learned sequentially and use the superscript $(T)$ to denote that a variable is associated with a task, for example: $\bm w^{(T)}$. The model is first trained on task $A$. Consider a new task $B$ with a new probe $\bm w^{(B)}$. We train our feature-reader model---with $\bm w$ and $\Phi$ as learnable parameters---through \textit{gradient descent} on task \(B\) using MSE with step size \(\eta\): 
\begin{equation}
\mathcal L_B(\bm x)=\tfrac12(\hat y^{(B)}(\bm x)-y^{(B)}(\bm x))^2.
\end{equation}
Before analyzing the gradients and loss, we introduce two quantities that enhance the interpretability of the results. 

\begin{definition}[Feature contribution $\beta_i$]
The \emph{feature contribution} measures how predictive feature $i$ is for the task:
\begin{equation}
    \beta_i^{(T)} := \mathbb{E}_{\bm x}[y^{(T)}(\bm x) f_i(\bm x)].
\end{equation}
\end{definition}
Positive values indicate activation for positive labels, and negative values for negative labels. This measure will prove useful to identify important features in practical models (\cref{sec:practical_case}).

% It would also be helpful to assess the alignment between the probe and a particular feature vector after convergence:

\begin{definition}[Probe-feature alignment $\gamma_i$]
The \emph{probe-feature alignment} measures how strongly the probe relies on feature $i$:
\begin{equation}\label{eq:probe_sensitivity}
    \gamma_i^{(T)} := (\bm w^{(T)})^\top \bm \phi_i.
\end{equation}
\end{definition}
A large $|\gamma_i|$ implies strong reliance on feature $i$, with the sign indicating whether it's used for positive or negative predictions. In \cref{sec:toy_experiments} and \cref{sec:practical_case} we will use $\gamma_i$ to track readout misalignment. %and fading. 
If $\gamma_i^{(T)}$ for a task $T$ changes its value after training on later tasks, it means that the probe has lost the ability to read the feature as it used to do.

By combining these definitions, we can express the expected gradient updates for both the features and the probe, revealing a symmetric co-adaptation during training of task $B$: 

\begin{theorem}[Gradient dynamics of features and probes]\label{theorem:updates}
Let the feature co-activation $\Sigma_{i,j}^{(T)}:=\mathbb{E}_{\bm x\sim\mathcal D_T}[f_i(\bm x)f_j(\bm x)]$. Under Task \(B\) training, the expected gradient updates for $\bm \phi$ and $\bm w$ are:
\begin{equation}\label{eq:updates}
\mathbb{E}_{\bm x\sim\mathcal D_B}[\Delta\bm\phi_{i}] = -\eta \left( \sum_{j}\gamma_{j}^{(B)}\Sigma_{ij}^{(B)} - \beta_{i}^{(B)} \right) \bm w^{(B)}, \qquad
\mathbb{E}_{\bm x\sim\mathcal D_B}[\Delta \bm w^{(B)}] = -\eta \sum_{i} \left( \sum_{j}\gamma_{j}^{(B)}\Sigma_{ij}^{(B)} - \beta_{i}^{(B)} \right) \bm \phi_{i}.
\end{equation}
\end{theorem}
\begin{proof}
    See \cref{apx:proof_updates}.
\end{proof}

% \Cref{eq:updates} reveals two key observations:
% \begin{enumerate}[label=(\roman*), topsep=0pt,itemsep=0ex,partopsep=1ex,parsep=1ex]
%     \item If an old feature $f_i$ is rarely present in task $B$ (i.e., ~$\Sigma_{i,j}^{(B)} \approx 0$ for all $j$ and $\beta_i^{(B)}\approx 0$), then, $\mathbb{E}_{\bm x\sim\mathcal D_B}[\Delta\bm \phi_i] \approx 0$. \textbf{The feature vector receives no gradient pressure, and it exerts no pull on the probe's update.}
%     \item The update direction for the feature vectors $\bm \phi_i$ lies in the direction of the probe $\bm w$, while the update direction for the probe lies in the direction of a weighted sum of the feature vectors. Together, this means \textbf{gradient descent naturally pulls the active feature vectors and the task probe toward one another}.
% \end{enumerate}

% These observations have several implications. In ideal scenarios, observation (i) would mean that \textbf{forgetting is constrained to the active subspace}, therefore:

\Cref{eq:updates} reveals that the update direction for $\bm \phi_i$ lies along $\bm w^{(B)}$, while the update for $\bm w^{(B)}$ lies along a weighted sum of feature vectors, meaning \textbf{gradient descent naturally pulls active feature vectors and the task probe toward one another}. The implications depend on which features are active in task $B$:

\begin{corollary}[Best-case scenario for forgetting]\label{cor:best_case}
If tasks A and B activate completely disjoint sets of features, then for any task A feature $i$, its co-activation ($\Sigma_{ij}^{(B)} = 0$) and contribution ($\beta_i^{(B)} = 0$) in task B are strictly zero. Consequently, under \cref{theorem:updates}, $\mathbb{E}_{\bm x\sim\mathcal D_B}[\Delta\bm\phi_i] = 0$. The feature vector receives no gradient pressure, shielding old features.
\end{corollary}

However, this no longer holds under regularized optimizers. For instance, \textbf{weight decay} introduces fading in inactive or weakly contributing features, \textbf{making some optimizer choices a potential risk factor for forgetting}. Conversely, the worst-case forgetting scenario occurs when tasks share features but assign them conflicting contributions:

\begin{corollary}[Worst-case forgetting scenario]\label{cor:worst_case}
    The update magnitude for both feature $\bm\phi_i$ and $\bm w^{(B)}$ is governed by the inner term $\sum_{j}\gamma_{j}^{(B)}\Sigma_{ij}^{(B)} - \beta_{i}^{(B)}$. This term is large when features are strongly co-activated ($\Sigma_{ij}^{(B)}$ large) with features aligning with the probe ($\gamma_j^{(B)}$ large) yet contribute conflictingly to the new task ($\beta_i^{(B)}$ opposite in sign to $\beta_i^{(A)}$).
\end{corollary}
\begin{remark}\label{remark:feature_uninformative}
    Another case worth noting occurs when the feature is uninformative for the new task ($\beta_i^{(B)} \approx 0$), rendering updates purposeless. While the scenario in Corollary \ref{cor:worst_case} alters feature geometry despite potentially preserving the feature's capacity, the latter case provides the network no incentive to maintain the feature's capacity.
\end{remark}
% \begin{corollary}[Worst-case forgetting scenario]\label{cor:worst_case}
% Under \cref{theorem:updates}, for a single dominant active feature $i$ with $\Sigma_{ij}^{(B)} \approx 0$ for $j \neq i$, the step size satisfies:
% \begin{equation}
%     \|\mathbb{E}_{\bm x\sim\mathcal D_B}[\Delta\bm\phi_i]\| = \eta \cdot |\gamma_i^{(B)}\Sigma_{ii}^{(B)} - \beta_i^{(B)}| \cdot \|\mathbf{w}^{(B)}\|
% \end{equation}
% \end{corollary}
% \begin{remark}
%     The worst case forgetting case can occur as an artifact of probe initialization and arises in two scenarios: (1) the new probe aligns with the previous one but the feature contribution reverses sign, $\beta_i^{(B)} = -\beta_i^{(A)}$; or (2) the new probe is antialigned with the previous one while the feature contribution is unchanged, $\beta_i^{(B)} \approx \beta_i^{(A)}$. Another case worth noting is when the feature is uninformative for the new task, $\beta_i^{(B)} \approx 0$, as the updates are then not purposeful for the task. % Given that randomly initialized vectors tend to be near orthogonal in high-dimensional spaces, this last case may be the most common in practice.
% \end{remark}

% Furthermore, \cref{eq:updates} implies that in the limit of infinite gradient updates, the set of active feature vectors will rotate towards the direction of the probe, \textbf{causing the rank of the representation to collapse}. This observation is reminiscent of the neural collapse phenomenon~\citep{papyan2020neural}. 

\paragraph{The effect of having multiple probes}
Up to this point, our analysis has focused on a single task probe. However, practical networks use multiple simultaneous probes. If we extend our framework to $p$ probes, the expected gradient update on a feature $\bm\phi_i$ becomes a sum over all individual probes $\bm w_k^{(B)}$:
\begin{equation}
    \mathbb{E}_{\bm x\sim\mathcal D_B}[\Delta\bm\phi_i] = -\eta \sum_{k=1}^p \left( \sum_j \gamma_{j,k}^{(B)}\Sigma_{ij}^{(B)} - \beta_{i,k}^{(B)} \right) \bm w_k^{(B)}
\end{equation}
This summation makes the geometric risk of forgetting explicit. With a single probe, an active feature with low contribution to the current task has more chances of staying in the probe's $(m-1)$-dimensional orthogonal null space. However, as the number of task probes $p$ increases, this orthogonal space shrinks to $m-p$ dimensions in the worst case. Features can no longer easily remain orthogonal to all probes, subjecting them to inescapable gradient pressure, which empirically manifests as capacity degradation as they get compressed towards the null-space (\cref{sec:results_readout}). We extend the analysis to cover the case where probes are used across tasks in \cref{apx:feature_updates_shared}, where we find a term that suppresses features predictive for task $A$.

\paragraph{A brief analysis on depth}
Finally, we briefly discuss the effect of depth. Unlike the single-layer case, in deeper networks, each feature vector at a given layer is implicitly determined by earlier computations, meaning that any update to a feature must be coordinated across multiple layers simultaneously. We conjecture that this interdependency is sensitive, and any transformations occurring at any layer may have compounding effects on features from later layers. When the circuitry of a feature is altered at any preceding layer, this translates to capacity degradation. Moreover, this coordination requirement may affect our two primitive transformations asymmetrically in worst-case settings: preserving a feature's direction would require all layers to jointly maintain consistent circuitry, while norm reduction can arise from independent updates at any single layer, making fading the more likely outcome under gradient pressure. This prediction is consistent with our empirical findings in \cref{sec:results_toy_depth}.

\subsection{Conclusions}
These results predict the best- and worst-case scenarios for forgetting: disjoint feature sets minimize the impact on features and forgetting, while overlapping active subspaces maximize it. Furthermore, our analysis provides a \emph{mechanistic interpretation of the empirically observed U-shaped relationship between task similarity and forgetting}, where intermediately similar tasks (shared but misaligned features) interfere more than either very similar or very dissimilar tasks (disjoint sets of active features), consistent with empirical observations~\citep{ramasesh2021anatomy, lee2021continual, hiratani2024disentangling}. Next we further examine whether these theoretical predictions hold empirically.

\section{Experiments with the feature-reader model}\label{sec:toy_experiments}

As a first test of our analytical results, we implemented the feature-reader model. The inputs $\bm{x} \in \mathbb{R}^n$ are synthetic vectors whose dimensions directly correspond to the $n$ features of the model, simulating the disentangled structure we assume of true underlying task features. We generated synthetic data for five sequential regression tasks, each defined by a contribution vector $\bm{\beta}_t$. The model has a shared one-layer encoder $\Phi \in  \mathbb{R}^{m \times n}$ (with $n$ features to an $m$-dimensional activation space) and five probes ($\bm{W}^{(T)} 
\in \mathbb{R}^{5 \times m}
$) per task. It was trained with Adam and MSE loss (full experimental details are in \cref{apx:experimental_setup_featurereader}). To analyze the mechanisms of performance drop, we computed the following metrics:
\begin{itemize}[topsep=0pt,itemsep=0ex,partopsep=0ex,parsep=1ex, wide, labelwidth=!, labelindent=0pt]
    \item \rebuttal{\textbf{Avg. MSE}: Measures the task performance for all seen tasks.}
    \item \textbf{Misalignment}: Tracks degradation in probe-feature alignment ($\gamma_{i}$), which translates to readout misalignment. This metric accounts for misalignment due to reading active features stronger (negative values) or weaker (positive values).
    % \item \textbf{Cap. Degradation}: We track decreases in the allocated capacity of task-specific features.
    \item \textbf{Fading}: We monitor fading by tracking a reduction in feature vector norms $\|\bm \phi_i\|$.
    \item \textbf{$\Delta$Overlap}: We monitor changes in overlap by tracking unit-normalized feature capacities (normalized capacity). This metric isolates feature overlap from fading, thereby disentangling the two sources of capacity degradation.
\end{itemize}

\begin{figure}[t]
    \centering
    \vspace{-6pt}
    \begin{subfigure}{1\columnwidth}
    \begin{subfigure}{0.245\columnwidth}
    \includegraphics[width=\linewidth]{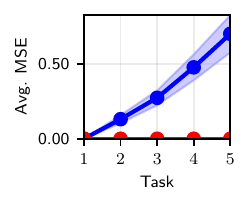}
    \end{subfigure}%
    % \hspace{4pt}
    \begin{subfigure}{0.245\columnwidth}
    \includegraphics[width=\linewidth]{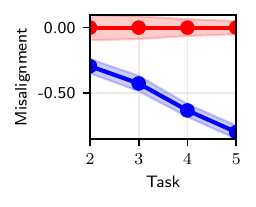}
    \end{subfigure}%
    \begin{subfigure}{0.245\columnwidth}
    \includegraphics[width=\linewidth]{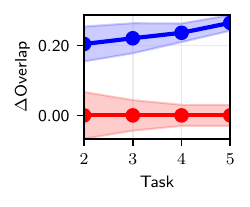}
    \end{subfigure}%
    % \hspace{4pt}
    % \hspace{0.25pt}
    \begin{subfigure}{0.245\columnwidth}
    \includegraphics[width=\linewidth]{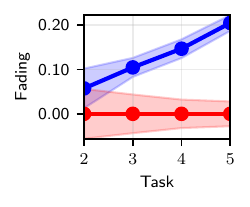}
    \end{subfigure}%
    \vspace{-19pt}
    \end{subfigure}
    
    \hspace{20pt}
    \includegraphics[height=26pt]{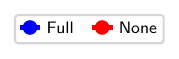}
    \vspace{-12pt}
    \caption{\textbf{Effect of shared features on capacity and MSE forgetting.} Tasks with shared but misaligned features (\texttt{Full}) show more forgetting (increase in Avg. MSE) due to stronger misalignment (which is negative, signaling that the probes read noisy features) and capacity degradation ($\Delta$Overlap and Fading) than tasks with disjoint sets of features (\texttt{None}).
    }
    \label{fig:shared_features}
\end{figure}
\vspace{-4pt}
\begin{figure}[t]
    \centering
    \vspace{-6pt}
    \begin{subfigure}{1\columnwidth}
    \begin{subfigure}{0.245\columnwidth}
    \includegraphics[width=\linewidth]{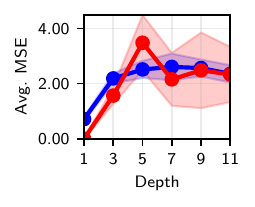}
    \end{subfigure}%
    % \hspace{4pt}
    \begin{subfigure}{0.245\columnwidth}
    \includegraphics[width=\linewidth]{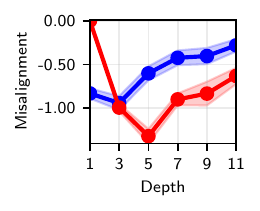}
    \end{subfigure}%
    \begin{subfigure}{0.245\columnwidth}
    \includegraphics[width=\linewidth]{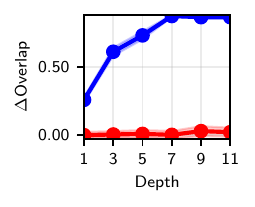}
    \end{subfigure}%
    % \hspace{4pt}
    % \hspace{0.25pt}
    \begin{subfigure}{0.245\columnwidth}
    \includegraphics[width=\linewidth]{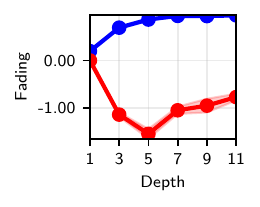}
    \end{subfigure}%
    \vspace{-19pt}
    \end{subfigure}
    
    \hspace{20pt}
    \includegraphics[height=26pt]{assets/toy_model_experiments/comparison_phi_depth_vs_full/legend_None.pdf}
    \vspace{-12pt}
    \caption{\textbf{Effect of depth on allocated capacity.} Shown are each metric at the end of training of the last task for different depths. Depth causes catastrophic capacity degradation in tasks with shared but misaligned features (\texttt{Full}). The effect is milder but still present for tasks with disjoint sets of active features (\texttt{None}).
    }
    \label{fig:effect_depth}
\end{figure}

\subsection{Results}

\paragraph{Worst- and best-case scenarios}

To test the predictions of \cref{sec:dynamics}, we vary feature coactivation and contribution across tasks. \Cref{fig:shared_features} shows two extremes: \texttt{Full}, where all features are coactive across tasks but only task-specific features have nonzero contribution, $\beta^{(B)}_i \approx 0$ for task $A$ features); and \texttt{None}, where tasks use disjoint feature sets ($\Sigma^{(B)}_{ii} \approx 0$), shielding task A features per \cref{cor:best_case}. 
%Following the noisy-feature case of \cref{cor:worst_case}, 
In line with our predictions, we found that \texttt{Full} produces the most forgetting: coactive but uninformative features ($\beta_i^{(B)} \approx 0$) receive no stabilizing gradient signal, causing capacity to degrade through increasing overlap and fading. Furthermore, the negative misalignment indicates that the probe is increasingly reading 
%noisy features rather than the task-relevant ones, 
task-irrelevant features, since misalignment is computed over all active features. For an extended analysis on task similarity beyond these two extreme cases, please refer to \cref{apx:extended_toy_experiments}.

\paragraph{Readout saturation amplifies forgetting and capacity degradation}\label{sec:results_readout}

\begin{figure}[t]
    \centering
    \vspace{-6pt}
    \begin{subfigure}{1\columnwidth}
    \begin{subfigure}{0.248\columnwidth}
    \includegraphics{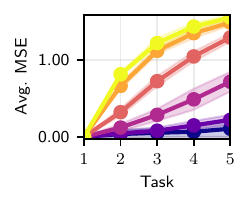}
    \end{subfigure}%
    % \hspace{4pt}
    \begin{subfigure}{0.248\columnwidth}
    \includegraphics{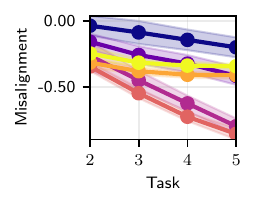}
    \end{subfigure}%
    \begin{subfigure}{0.248\columnwidth}
    \includegraphics{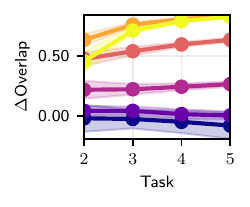}
    \end{subfigure}%
    % \hspace{4pt}
    % \hspace{0.25pt}
    \begin{subfigure}{0.248\columnwidth}
    \includegraphics{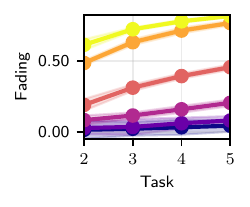}
    \end{subfigure}%
    \vspace{-8pt}
    \end{subfigure}
    \begin{subfigure}{1\columnwidth}
        \centering
        \makebox[\textwidth][c]{
            % \hspace{-1.5\columnwidth}
            \begin{minipage}{42pt}
                \flushright
                \small Nº Probes:
            \end{minipage}%
            \begin{minipage}{180pt}
                \includegraphics{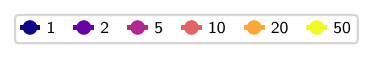}
            \end{minipage}%
        }
    \end{subfigure}
    \vspace{-10pt}
    \caption{\textbf{Impact of the number of probes on allocated capacity.} Increasing the number of probes that read the activation leads to increased forgetting as previous features cannot escape gradient pressure, causing capacity degradation (overlap and fading).}\label{fig:readout_saturation}

\end{figure}

We next examine how the volume of activation space covered by readouts affects forgetting. We focused on the \texttt{Full} scenario, since \texttt{None} remains unaffected (\cref{cor:best_case}). In \cref{sec:dynamics}, we predicted that increasing the number of probes reduces the chance that any feature remains orthogonal to all probes, thereby limiting its invulnerability to the gradient signal. Consistent with this analysis, we found in \cref{fig:readout_saturation} that expanding the number of probes amplifies capacity degradation and leads to forgetting.

\paragraph{Depth also amplifies forgetting and capacity degradation}\label{sec:results_toy_depth}

In line with prior work~\citep{guha2024diminishing, lu2024revisiting}, we found that network depth exacerbates forgetting. To assess this effect and scale up to realistic models, we extended the encoder of the model, which parameterized the feature vectors, with additional layers. To isolate the effects of depth, we kept the model's linear expressiveness by not adding nonlinearities. We analyze the direction of each feature $i$ in activation space using $e_i^\top \Phi$, where $e_i$ is the $i$-th standard basis vector. In \cref{fig:effect_depth}, we found that depth impacts \texttt{Full} severely and even affects \texttt{None}. In the worst-case (\texttt{Full}), deeper encoders obliterate feature capacity, with persistent negative misalignment likely driven by previous probes reading spurious features. Interestingly, \texttt{None} (best-case) displays low capacity degradation and negative fading (increased feature norms), yet still suffers forgetting due to seemingly norm-induced misalignment.

\section{Identifying features with crosscoders}\label{sec:scalingup}

\begin{figure}[t]
    \centering
    % Left side: Crosscoder diagram
    \begin{minipage}{0.48\textwidth}
        \centering
        \includegraphics[height=90pt]{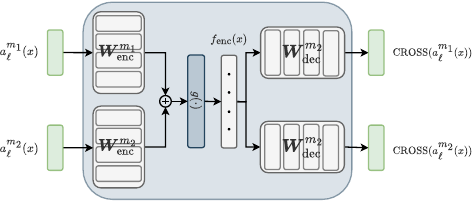}
        \caption{\textbf{Crosscoder diagram:} Representations $\bm a_{\ell}^{m_t}(\bm x)$ from multiple models are mapped to a single shared latent space $\bm f_{\text{enc}}(\bm x)$, which is then used by model-specific decoders to reconstruct the original activations.}
        \label{fig:crosscoder_diagram}
    \end{minipage}
    \hfill % Adds horizontal space between the two minipages
    % Right side: Activation maps
    \begin{minipage}{0.48\textwidth}
        \centering
        \includegraphics[height=90pt]{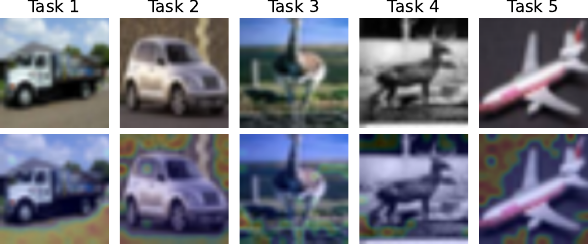}
        \caption{\textbf{Activation maps for the most important feature from task 1.} Shown are the images for which this feature showed the strongest activations for every task. The feature appears to correlate with background areas.}
        \label{fig:task_1_feature_178}
    \end{minipage}
    % \vspace{-4pt}
\end{figure}

Our feature-reader model provided key mechanistic insights, such as the detrimental effect of depth, using a simplified setting where the features were known. However, in deep neural networks, the features are not known beforehand. Thus, to apply our framework, we must first identify the features. In this section, we discuss tools that allow us to approximate and track these features in practical models. % This section introduces crosscoders~\citep{lindsey2025crosscoders}---a tool based on Sparse Autoencoders (SAEs)~\citep{bricken2023monosemanticity,huben2024sparse}---to identify and track these features in practical models.

%Under the linear representation hypothesis, features are linearly expressed, while the superposition hypothesis states that their number exceeds the neuron count ($d_{\text{model}}$), making them entangled. 
\textbf{Sparse autoencoders (SAEs)}~\citep{bricken2023monosemanticity} separate and identify features in superposition by mapping layer activations $\bm a_\ell(\bm x)\in\mathbb{R}^{d_{\text{model}}}$ to a higher-dimensional sparse space ($d_{\text{cross}}>d_{\text{model}}$):
\begin{equation}\label{eq:standard_sae}
    \bm f_{\text{enc}}(\bm x) = g(\mW_{\text{enc}}\bm a_\ell(\bm x) + \bm b_{\text{enc}}),\qquad \text{SAE}(\bm x) = \mW_{\text{dec}} f_{\text{enc}}(\bm x) + \bm b_{\text{dec}}
\end{equation}
Each latent unit $f_{\text{enc},i}(\bm x)$ corresponds to a feature activation, and each decoder column $\mW_{\text{dec},i}$ represents its contribution to the reconstructed activation.

\textbf{Crosscoders}~\citep{lindsey2025crosscoders} (\cref{fig:crosscoder_diagram}, see \cref{apx:introduction_crosscoders} for a detailed introduction) generalize SAEs to compare models or layers by learning a \emph{shared latent feature space} across representations ${\bm a_\ell^{m_t}(\bm x)}$ from multiple models $m_t\in\mathcal{M}$ while model-specific decoders $\mW_{\text{dec}}^{m_t}$ reconstruct each model’s activations:
\begin{equation}
\bm f_{\text{enc}}(\bm x) = g\Bigl( \sum_{{m_t} \in \mathcal{M}} \mW_{\text{enc}}^{m_t}
\bm a_{\ell}^{m_t}(\bm x) + \bm b_{\text{enc}}\Bigr), \qquad \text{CROSS}^{m_t}(\bm a^{m_t}_\ell(\bm x)) = \mW^{m_t}_{\text{dec}}\bm f_{\text{enc}}(\bm x) + \bm b_{\text{dec}}^{m_t}.
\end{equation}
This formulation embeds our conceptual framework (\Cref{sec:conceptualization}) in practical models. Note, however, that crosscoder features should not be interpreted as ground-truth. We treat the global feature basis defined by the encoders $\bm f_{\text{enc}}(\bm x)$ as an approximation of the features learned by the neural network $\bm f(\bm x)$, and each decoder column $\mW_{\text{dec},i}^{m_t}$ as a feature vector $\bm \phi_i^{(t)}$. This enables approximate tracking of features across training stages as exemplified in \cref{fig:task_1_feature_178}.

% the shared encoder defines a global feature basis corresponding to $\bm f(x)$, and each decoder column $\mW_{\text{dec},i}^{m_t}$ represents the feature vector $\bm \phi_i^{(t)}$ of model~$m_t$. Tracking these vectors across tasks reveals how features transform and overlap, empirically connecting mechanistic feature geometry with real network representations.

\begin{figure}[t]
    \centering
    % ================= Left Column (First Figure) =================
    \begin{minipage}[t]{0.49\textwidth}
        \centering  
        \vspace{-6pt}
        \begin{subfigure}{1\columnwidth}
            \begin{subfigure}{0.495\columnwidth}
            \includegraphics[width=\linewidth]{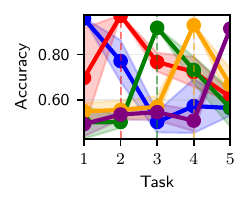}
            \end{subfigure}
            \begin{subfigure}{0.495\columnwidth}
            \includegraphics[width=\linewidth]{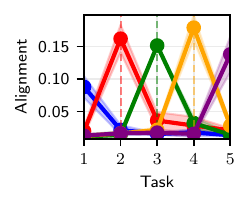}
            \end{subfigure}
        \end{subfigure}
        
        \vspace{-20pt}    
        
        \begin{subfigure}{1\columnwidth}
            \begin{subfigure}{0.495\columnwidth}
            \includegraphics[width=\linewidth]{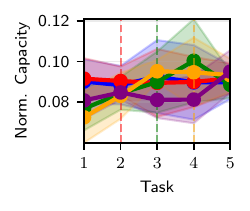}
            \end{subfigure}
            \begin{subfigure}{0.495\columnwidth}
            \includegraphics[width=\linewidth]{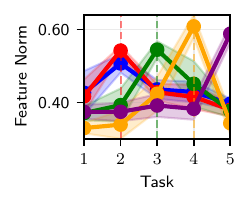}
            \end{subfigure}
        \end{subfigure}
        
        \vspace{-20pt}
        
        \begin{subfigure}{1\columnwidth}
            \begin{subfigure}{0.495\columnwidth}
            \includegraphics[width=\linewidth]{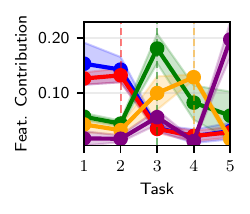}
            \end{subfigure}
            \begin{subfigure}{0.495\columnwidth}
            \includegraphics[width=\linewidth]{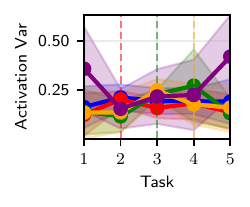}
            \end{subfigure}
            \vspace{-19pt}
        \end{subfigure}

        \begin{subfigure}{1\columnwidth}
        \makebox[\textwidth][c]{
            % \hspace{-1.5\columnwidth}
            \begin{minipage}{0.3\columnwidth}
                \flushright
                \small Metrics for task:
            \end{minipage}%
            \begin{minipage}{0.6\columnwidth}
                \includegraphics[height=26pt]{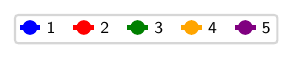}
            \end{minipage}%
        }
        \end{subfigure}
        \vspace{-10pt}
        \caption{\textbf{Metrics evolution across tasks for the penultimate layer.} Shown is the average value for each metric over the top-5 most important features of each task.
        }
        \label{fig:results_crosscoder}
    \end{minipage}
    \hfill % Adds space between the two columns
    % ================= Right Column (Second & Third Figures) =================
    \begin{minipage}[t]{0.49\textwidth}
        \centering
        
        % --- Top Figure (Originally Figure 2) ---
        \vspace{-6pt}
        \begin{subfigure}{1\columnwidth}
            \hspace{-8pt}
            \begin{subfigure}{0.495\columnwidth}
                \includegraphics[width=\linewidth]{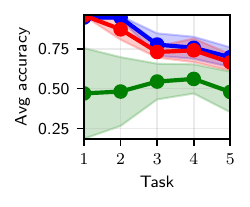}
            \end{subfigure}
            \hspace{2pt}
            \begin{subfigure}{0.495\columnwidth}
                \includegraphics[width=\linewidth]{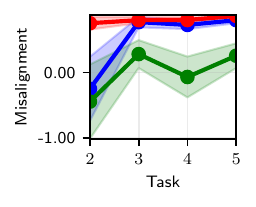}
            \end{subfigure}
        \end{subfigure}
        \begin{subfigure}{1\columnwidth}
        \vspace{-18pt}
        \centering
        \hspace{24pt}
        \includegraphics[height=26pt]{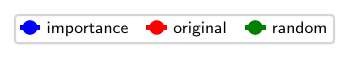}
        \vspace{-10pt}
        \end{subfigure}
        \caption{\textbf{Comparison of intervened classifier and baselines.} The intervened classifier mitigates misalignment, suggesting that fading dominates forgetting in this setup.}
        \label{fig:clf_interv_comparison}
        
        \vspace{15pt} % Adds vertical padding between the two stacked figures
        
        % --- Bottom Figure (Originally Figure 3) ---
        % \vspace{-4pt}
        \begin{subfigure}{1\columnwidth}
            \vspace{-12pt}
            \hspace{-8pt}
            \begin{subfigure}{0.495\columnwidth}
            \includegraphics[width=\linewidth]{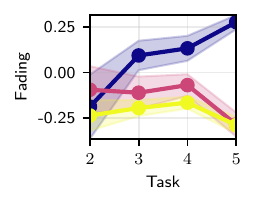}
            \end{subfigure}
            \hspace{2pt}
            \begin{subfigure}{0.495\columnwidth}
                \includegraphics[width=\linewidth]{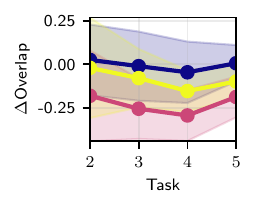}
            \end{subfigure}
        \end{subfigure}
        % \hspace{24pt}
        \begin{subfigure}{1\columnwidth}
            \centering
            \vspace{-10pt}   
            \hspace{28pt}
            \includegraphics[height=26pt]{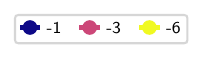}
        \vspace{-10pt}
        \end{subfigure}
        \caption{\textbf{Capacity metrics for different layers of the ViT.} While the last layer feature's capacities degrade, earlier layers grow theirs.}
        \label{fig:results_capacity_layers}
        
    \end{minipage}
\end{figure}

\subsection{An example of a case study with a ViT}\label{sec:practical_case}

% \begin{figure}[t]
%   \centering
%   \begin{subfigure}{1\columnwidth}
%   \centering
%   \includegraphics[width=0.9\columnwidth]{assets/crosscoders/results/example_grid.pdf}
%   \end{subfigure}
%   \caption{\textbf{Activation maps for the most important feature from task 1.}
%   Shown are the images that most strongly activated this feature at each task. The bottom row overlays these images with their corresponding activation maps. The feature appears to be activated by the background parts of the image.
%   }
%   \label{fig:task_1_feature_178}
% \end{figure}

To demonstrate how real models can be studied under our framework, we show how one can apply crosscoders to study a Vision Transformer (ViT) sequentially trained on CIFAR-10. While our tractable model analyzed the direct gradient dynamics of known feature vectors $\Phi$, this ViT analysis tracks their analogs---the crosscoder's decoder vectors ($\mW_{dec,i}^{m_t}$)---to study the same geometric transformations. 

\paragraph{Setup} We train a ViT~\citep{dosovitskiy2021an} on Split CIFAR-10~\citep{krizhevsky2009learning} (five tasks, two classes per task), under a task-incremental learning setup~\citep{van2022three} (one head per task) across four seeds. We train a separate crosscoder for each seed. For each task $t$, we analyze its model $m_t$ by selecting the top five features ranked by importance, $I^{(t)}_i=\beta^{(t)}_i \cdot \gamma^{(t)}_i$. This product captures features that are both highly predictive for the task (high Feat. Contribution $\beta^{(t)}_i$) and strongly relied upon by the model's readout (high Alignment $\gamma^{(t)}_i$). % We empirically found that \textbf{$\beta^{(t)}_i$ alone serves as a good proxy of a feature's importance} (as seen in the bottom-left plot of \cref{fig:results_crosscoder}), making it a particularly easy-to-compute importance measure. 
Full setup details are in \cref{apx:crosscoders_experimental_details}.

\paragraph{Identifying fading and misalignment as the drivers of forgetting in a task-incremental setup}

For this example experiment, \cref{fig:results_crosscoder} shows that each task's accuracy drops as new tasks are learned, despite a largely stable normalized capacity. This suggests that severe overlap between features is not the main issue. Instead, the observed metric changes are more consistent with fading and misalignment than with feature overlap in this task-incremental setting (see \cref{apx:class-incremental} for the class-incremental variant, which shows a different pattern). A notable exception occurs after the second task, where norms increase, likely due to the strong similarity and shared features between the first and second tasks, as indicated by their closely aligned contribution $\bm \beta$ values. Note that all features remain active across tasks (bottom-right plot), but their contribution (bottom-left plot) changes substantially. This seems to be consistent with the \emph{worst-case scenario} described in our theory: active but misaligned features receive a high gradient pressure. % The observed dominance of fading over rotation aligns with trends seen in the feature-reader model under increasing depth (\cref{sec:results_toy_depth}). 
 % Maybe verify with visualizations? If enough space...

\paragraph{Testing misalignment as the culprit}\label{sec:testing_misalignment}

As an example of how we could intervene the network, we performed an intervention to isolate misalignment in which we crafted a classifier head by linearly combining the evolved decoder vectors $\mW^{m_t}_{\text{dec},i}$ weighted by their initial importance $I^{(t)}_i$. This would restore the original readout alignment and, if there were no capacity degradation, the performance should be recovered. We compare against the \texttt{original} classifier and a \texttt{random} combination of decoder vectors. As shown in \cref{fig:clf_interv_comparison}, accuracy was recovered only for task 1 (which shares features with task 2), and only marginally for the others, suggesting that fading is the dominant factor in performance decay in this task-incremental case study. The high forgetting observed in the alignment $\bm\gamma$ is not evidence of rotational misalignment but rather a direct result of this fading (norm reduction). \Cref{fig:results_crosscoder} is consistent with this: while the norms of key features increase during training on the task, they subsequently decrease as the model reallocates capacity to newer tasks.

\paragraph{Studying earlier layers}

% \begin{figure}[t]
%     \centering
%     \vspace{-6pt}
%     \begin{subfigure}{1\columnwidth}
%     \hspace{-6pt}\includegraphics{assets/crosscoders/results/crosscoder_seq1_forgetting/norms_forgetting_evolution_by_task_plasma.pdf}
%     \begin{subfigure}{0.45\columnwidth}
%     \end{subfigure}
%     \hspace{-8pt}
%     \begin{subfigure}{0.45\columnwidth}
%     \includegraphics{assets/crosscoders/results/crosscoder_seq1_forgetting/capacities_norm_forgetting_evolution_by_task_plasma.pdf}
%     \end{subfigure}
%     \vspace{-20pt}
%     \end{subfigure}
%     \vspace{-8pt}
%     \hspace{32pt}
%     \includegraphics{assets/crosscoders/results/crosscoder_seq1_forgetting/legend_plasma.pdf}
%     \caption{\textbf{Capacity metrics for different layers of the ViT.} While the last layer's capacity degrades, earlier layers grow theirs.
%     }
%     \label{fig:results_capacity_layers}
% \end{figure}

A natural next question could be whether this pattern holds across layers. We studied two earlier layers in \cref{fig:results_capacity_layers} and found that earlier layers \emph{gained} allocated capacity as new tasks were learned. Feature norms increased in these layers. Even though we are showing a limited-scale experiment, unlike prior work~\citep{yosinski2014transferable, ramasesh2021anatomy} reporting that earlier layers remain stable while forgetting mainly arises in the final layers, these results tentatively suggest they may actively gain capacity, a discrepancy worth investigating further.

\rebuttal{In this section, we have presented a case study under a task-incremental setting. In \cref{apx:class-incremental}, we also provide an additional case study under a class-incremental setting, which displays higher overlap and a slightly different behaviour.}

\section{Discussion and limitations}

% This framework might help us better understand different phenomena in continual learning such as the stability gap.
% OK desde el punto de vista de mechinterp, se podría mirar algo más de circuit-level. Cómo interaccionan los diferentes model components con el forgetting 
% también comparar diferentes métodos de continual learning
% identification of important knowledge that we want to preserve or forget
% One promising direction we envision would allow practitioners to identify concrete, relevant features that should be preserved, or, on the other hand, useless or confounding features that should be forgotten to make room for new ones.

% This paragraph feels too generic and like a conclusion right now, should rewrite
% I want to write something about depth and what "overwriting" means when people say that previous knowledge is overwritten when a new task is learned. It might not be as simple as that and this framework gives a concrete interpretation of what overwriting really means.
We have bridged continual learning and mechanistic interpretability, which has resulted in a conceptual framework that provides a granular vocabulary for forgetting. We have studied a toy model that we refer to as the feature-reader model, which functions as a proxy for analyzing the mechanisms of feature forgetting in practical neural networks. Similar toy models have proven valuable in prior work~\citep{elhage2022toy, hanni2024mathematical} in unveiling phenomena such as superposition. In this paper, we used the framework to gain a better mechanistic understanding of catastrophic forgetting through mechanistic quantities that complement existing techniques, such as CKA or linear probing (\cref{apx:extended_toy_experiments}). \rebuttal{In future work, the framework could be used to track our metrics on more fine-grained timescales, e.g., around task transitions to study their behaviour during the stability gap~\citep{lange2023continual}.}

We discussed that capacity degradation leads to forgetting that cannot be recovered by later layers. There may be other cases where knowledge is irrecoverable as well. Although simple readout misalignment can, in principle, be corrected by adapting the readouts, this recoverability may be limited in capacity-constrained settings with overlapping features. Models may organize such features to reduce interference~\citep{elhage2022toy}. As suggested by the joint behavior of probe accuracy and capacity metrics in \cref{apx:extended_toy_experiments}, conflicting updates from new tasks can induce geometric rearrangements that disrupt this organization, increasing unavoidable interference in the readout. In such cases, performance degradation may no longer be recoverable by readout realignment alone.

As is common in the mechanistic literature, our work relies on the linear representation hypothesis. Although recent work shows that some features, such as days of the week~\citep{engels2024not}, are multidimensional, substantial evidence supports the hypothesis in diverse contexts~\citep{arora2018linear, o2023disentangling, bricken2023monosemanticity}. We also assume that the hypothesis remains valid during continual learning. Yet, when a feature fades, it could also be questioned whether the feature is still linear. 

As an example of how our framework can be applied to practical models, we studied feature dynamics in a ViT continually trained on Split CIFAR-10. While the results revealed fading predominance and layer-wise capacity dynamics under task-incremental training, they should be considered preliminary given the limited scale of this demonstration. Also, we caution that SAEs are empirical tools with limitations. They are not foolproof and can be difficult to tune, as feature decompositions are approximate and sensitive to modeling choices~\citep{gao2025scaling}. \rebuttal{Some relevant challenges include feature consistency (the SAE captures different features with different seeds)~\citep{song2025positionmechanisticinterpretabilityprioritize} and feature split or absorption~\citep{bussmann2025learning}. It remains to be tested how sensitive our conclusions for our results in \cref{sec:practical_case} are to variability in the crosscoder.}

\textbf{Implications for continual learning algorithms} \hspace{1em}
Our results suggest that mitigating forgetting requires explicitly preserving the feature-level circuitry across layers, rather than only constraining parameter drift or final representations. The detrimental effect of depth indicates that feature preservation is a coordinated, multi-layer problem, pointing toward strategies that stabilize feature geometry throughout the network. For instance, by maintaining alignment between feature vectors and their downstream readouts, or by protecting high-contribution features from destructive updates. From this perspective, we conjecture that architectures such as Mixture-of-Experts~\citep{litheory} may be beneficial by enforcing feature isolation while permitting reusage. Finally, our framework suggests a more proactive role for sparse autoencoders. \rebuttal{We envision a line of work that explores training sparse autoencoders concurrently with the target model, rather than as a post-hoc analysis step. This would enable real-time feature tracking during training, opening the door for active knowledge preservation and "smart" structural rearrangements. This direction would reframe continual learning as the problem of maintaining a stable, interpretable feature basis and its associated circuitry over time.}

% Despite these limitations, our framework has its own merits. While diagnostic tools such as linear probing or centered kernel alignment (CKA) are valuable for detecting representational drift, they often do not distinguish between the specific manifestations identified in this work: overlap, fading, and readout misalignment. For example, a drop in CKA signals a change but does not specify if the features have rotated (requiring orthogonalization) or faded (requiring norm preservation). Our framework addresses this gap, offering practitioners a semantic lens to diagnose future methods and, e.g., determine what knowledge to retain or forget.

% Implications for algorithms:
% - Focus on keeping alignment across layers. Might explain some properties on MoEs in continual learning "The broad takeaway is that MoE helps most when it is used as a mechanism for isolation and specialization, not just sparsity. If routing is unstable, or if the representation before routing is already entangled, MoE can still forget a lot." (https://www.perplexity.ai/search/are-there-any-works-studying-f-3OrRiGxcSOmM9LlFqAFY1w). 
% - If we could train SAEs continuously to track features during continual learning, they could become an active part of the feature-preservation process. We would just need to track the contribution of each feature and protect the important ones. SAE as part of the architecture.

\section{Conclusions}
% In this paper, we have introduced a conceptual framework that bridges continual learning and mechanistic interpretability. Using this framework, we have described best- and worst-case scenarios for forgetting. Non-active features tend to be preserved, while active but misaligned features risk degrading their representation or becoming misaligned with downstream computations. We have formalized these intuitions with a tractable feature-reader model, which has allowed us to derive bounds on forgetting. Empirical evidence with a continual learning simulation of this model has supported the formal analysis and has revealed the detrimental effect of depth on forgetting and allocated capacity, particularly on the norm of features. Finally, we have studied a ViT with crosscoders. This analysis has revealed how fading at the last layer seems to be the main responsible for forgetting.

We have introduced a new, feature-centric framework that bridges continual learning and mechanistic interpretability. We showed how forgetting can be interpreted as the result of geometric transformations that reduce the allocated capacity of features or cause readout misalignment. We characterized the best and worst scenarios for forgetting and formalized this in a tractable feature-reader model. Empirical validation of this model also revealed the detrimental effect of depth. Finally, we have demonstrated how this analysis can be carried out in practical models through the use of crosscoders. We studied a ViT trained sequentially and identified the mechanisms of forgetting in this controlled setting. This work provides a new vocabulary for deepening the understanding of forgetting.

% \section*{Impact statement}

% This paper presents work whose goal is to advance the field of machine learning. There are many potential societal consequences of our work, none of which we feel must be specifically highlighted here.

\section*{Acknowledgements}
We want to thank Pau Rodríguez and Timm Hess for their valuable feedback and advice. We also appreciate the rich discussions with Jonathan Swinnen, Milan van Maldegem, Zehao Wang, Minye Wu, and Matthew Blaschko.

This paper has received funding from the Flemish Government under the Methusalem Funding Scheme (grant agreement n° METH/24/009).

{
    \small
    \bibliographystyle{ieeenat_fullname}
    \bibliography{main}
}

% \bibliographystyle{iclr2026_conference}
% \bibliography{main}

\clearpage
\appendix
\onecolumn
\section{Comprehensive related work}\label{apx:related_work}

\subsection{Theoretical and conceptual understanding of catastrophic forgetting}

Catastrophic forgetting, first identified decades ago \citep{MCCLOSKEY1989109, Ratcliff1990ConnectionistMO}, remains a central challenge in continual learning. While many methods have been proposed to mitigate it, a complete theoretical account is still developing. Early intuitions framed forgetting with high-level concepts like "weight drift" or "interference", but these lack the precision needed for a deep, mechanistic understanding.

Recent theoretical work has sought to add rigor from several angles:

\paragraph{Formalisms, bounds, and convergence} Earlier works formalized the continual learning problem and its scenarios~\citep{van2022three} and showed that continual learning is NP-hard by relating it to a set intersection decision problem~\citep{knoblauch2020optimal}. Recently, a significant body of work uses simplified settings, particularly linear models, to achieve analytical tractability. These studies have provided valuable insights, including deriving performance bounds \citep{evron2022catastrophic, evron2023continual, ding2024understanding}, proving convergence to joint solutions in cyclic learning \citep{jung2025convergence}, and proving global convergence in continual settings~\citep{zhu2025global}. In line with these works, we also employ a tractable linear model. However, where much of this work focuses on proving convergence or bounding final performance, our goal is to reveal the internal geometric dynamics of how representations are transformed and evolve during continual learning.

\paragraph{Task similarity and gradient alignment} Another prominent line of research investigates how task relationships influence forgetting. Forgetting is often framed as a consequence of misaligned task gradients or competition for representational subspaces. Studies have shown a U-shaped relationship in which intermediate task similarity can cause the greatest interference \citep{ramasesh2021anatomy, lee2021continual, hiratani2024disentangling}. Other work operating in the neural tangent kernel regime has linked forgetting to the parameter-space alignment of task updates \citep{bennani2020generalisation, doan2021theoretical}. In a recent publication, \citet{wanunderstanding} claimed that the cosine similarity between task "signal vectors" determines the degree of forgetting, with anti-aligned tasks being the most detrimental. 

These works successfully predict that certain task configurations cause forgetting, but they do not explain the representational mechanisms by which this forgetting occurs. Task-level metrics (vector alignment, subspace overlap) describe the conditions for interference but remain agnostic to the specific geometric transformations and capacity reallocation that degrade individual features. Our framework complements these predictive models by providing a mechanistic account of the internal dynamics they summarize. We decompose these high-level task conflicts into interpretable geometric transformations of the underlying feature vectors, directly linking them to changes in capacity and readout.

Taking a step toward these representational mechanisms, recent work~\citep{li2025towards} studies the dynamics of forgetting in a two-layer convolutional neural network. Using a data model of mutually orthogonal features, they find that forgetting occurs when a task-specific feature learned in the first task reappears in the second task with a strong signal but no label correlation. This mechanism aligns with our worst-case scenario (\cref{cor:worst_case} and \cref{remark:feature_uninformative}), where active but uninformative features undergo destructive updates. However, their assumption of perfect feature orthogonality does not consider feature capacity or superposition. Our framework addresses these by explicitly modeling feature geometry and distinguishing between feature allocated capacity and downstream readout alignment.

\paragraph{Architectural factors} The role of model architecture is also a key area of study. Overparameterization, particularly width, has been shown to mitigate forgetting, often by enabling a "lazy" training regime where parameters stay close to their initialization \citep{pmlr-v206-goldfarb23a, Goldfarb2025ARXIV_Analysis_of_Overparameterization, mirzadeh2022wide, guha2024diminishing, lu2024revisiting}. Our mechanistic framework could complement these findings by offering a potential explanation for how more capacity (i.e., more dimensions) can reduce the destructive feature overlap that causes forgetting, particularly by allowing more room for features to stay orthogonal to any readout.

\rebuttal{\paragraph{Representation and readout forgetting} A growing body of literature studies forgetting at the level of representations and forgetting at the readout or classification head.  \citet{hess2024knowledge} show that, while knowledge seems to accumulate over time in the representations, forgetting still occurs at the representation level. Through the use of CKA, \citet{ramasesh2021anatomy} identify that forgetting concentrates in the deeper layers of the model. Recent work~\citep{lanzillotta2026heads} studies forgetting at the representation level and at readout from the neural collapse lens. Their analysis suggests that minimal replay mitigates representation forgetting, while the readout requires more replay. Some task similarity works also consider readout similarity~\citep{hiratani2024disentangling, lee2021continual}. However, most of these works, except \citet{ramasesh2021anatomy}, focus their analysis on the penultimate layer of the neural network. Our framework applies to any layer of a neural network and is not restricted to the last readout (normally, the classifier head). Moreover, our work provides a fine-grained, mechanistic view into representation forgetting.}

\subsection{Mechanistic interpretability}

To build our feature-level account of forgetting, we draw on tools and concepts from the field of mechanistic interpretability, which aims to reverse-engineer the internal computations of neural networks.

\paragraph{Features, polysemanticity, and superposition} 
A central finding in this field is that knowledge is not typically stored in individual neurons. Instead, neurons are often polysemantic, participating in the representation of multiple concepts~\citep{mu2020compositional}. The fundamental unit of representation is therefore considered the feature, which is best understood as a direction in a layer's activation space~\citep{olah2020zoom, elhage2022toy}. Certain architectural choices or conditions, such as specific non-linearities allow a model to learn more features than it has neurons (or dimensions). This phenomenon is known as superposition. Superposition may be possible following the Johnson–Lindenstrauss lemma, which states that $exp(m)$ vectors can be placed almost-orthogonally in an $m$-dimensional space. Recent evidence points to the existence of superposition in deep neural networks, including large language models~\cite{gurnee2023finding, templeton2024scaling}.

\paragraph{Capacity and interference} 
Superposition has a direct cost: features must compete for finite representational resources. The geometry of the feature vectors---their norms and pairwise inner products---determines the allocated capacity for each feature, a measure of how cleanly it can be distinguished from others~\citep{elhage2022toy, scherlis2022polysemanticity}. This concept is crucial for our work. Pioneering research has suggested that superposition is a primary cause of interference in related problems like lifelong model editing~\citep{hu2024knowledge}, and concurrent work has empirically studied forgetting in LLMs using mechanistic insights~\citep{imanov2026mechanisticanalysiscatastrophicforgetting}.
%to our knowledge, we are the first to explicitly link the mechanistic concepts of superposition and allocated capacity to catastrophic forgetting in continual learning.

\paragraph{Linear dynamics analysis}
\Cref{theorem:updates} reinterprets the linear network dynamics of \citet{saxe2014exact} through the mechanistic lens of superposition~\citep{elhage2022toy}. \citet{saxe2014exact} decomposed the input-output covariance matrix ($\Sigma^{31}$ in their work) via SVD and described the dynamics of "connectivity modes"; The number of modes is limited to the rank of $\Sigma^{31}$. Superposition, however, creates a "congested" representation that an SVD-based analysis cannot fully capture. % Our framework explicitly models superposition. 
% Maybe cite work where they also study the gradient wrt features?

% \newline
% \newline
Our work explicitly links the mechanistic concepts of superposition and allocated capacity to catastrophic forgetting in continual learning. By synthesizing these two fields, our work provides a new lens through which to understand the geometric and representational roots of catastrophic forgetting. We move beyond describing \emph {that} forgetting happens and aim to explain precisely \emph {how} it happens at the level of the model's fundamental building blocks: its features.

\section{Proof of \cref{theorem:updates}}\label{apx:proof_updates}
\begin{proof}
We start from the per-example squared loss on task \(B\),
\[
\mathcal L_B(x)=\tfrac12\big(\hat y^{(B)}(x)-y^{(B)}(x)\big)^2,
\qquad
\hat y^{(B)}(x)=w^{(B)\top}a_\ell(x).
\]
First, by the chain rule for \(\phi_i\) (recall \(\bm{a}_\ell(\bm{x}) =\sum_{i=1}^{n_\ell} f_{\ell,i}(\bm{x}) \, \bm\phi_{\ell,i}.\)),
\[
\nabla_{\phi_i}\mathcal L_B(x)
= (\hat y^{(B)}(x)-y^{(B)}(x))\,\nabla_{\phi_i}\hat y^{(B)}(x)
= (\hat y^{(B)}(x)-y^{(B)}(x))\, w^{(B)} f_i(x).
\]
Next, expand \(\hat y^{(B)}(x)\) using the feature decomposition:
\[
\hat y^{(B)}(x)=w^{(B)\top}a_\ell(x)
= \sum_j (w^{(B)\top}\phi_j)\, f_j(x).
\]
Substitute this into the gradient expression:
\begin{align*}
\nabla_{\phi_i}\mathcal L_B(x)
&= \Bigg(\sum_j (w^{(B)\top}\phi_j)\, f_j(x) - y^{(B)}(x)\Bigg) w^{(B)} f_i(x).
\end{align*}

Taking expectation over $x \sim \mathcal D_B$ and using linearity,
\begin{align*}
\mathbb{E}_{\mathcal D_B}[\nabla_{\phi_i}\mathcal L_B]
&= \left( \sum_j (w^{(B)\top}\phi_j)\, \mathbb{E}[f_j(x) f_i(x)] - \mathbb{E}[y^{(B)}(x) f_i(x)] \right) w^{(B)}.
\end{align*}
Substitute the definitions
\(\gamma_j^{(B)}=w^{(B)\top}\phi_j\), \(\Sigma_{ij}^{(B)}=\mathbb{E}[f_i f_j]\), and \(\beta_i^{(B)}=\mathbb{E}[y^{(B)} f_i]\) to obtain
\[
\mathbb{E}_{\mathcal D_B}[\nabla_{\phi_i}\mathcal L_B]
= \Bigg(\sum_j \gamma_j^{(B)}\Sigma_{ij}^{(B)} - \beta_i^{(B)}\Bigg)\, w^{(B)}.
\]
Similarly, for the probe \(w^{(B)}\), we apply the chain rule:
\[
\nabla_{w^{(B)}}\mathcal L_B(x)
= (\hat y^{(B)}(x)-y^{(B)}(x))\,\nabla_{w^{(B)}}\hat y^{(B)}(x)
= (\hat y^{(B)}(x)-y^{(B)}(x))\, a_\ell(x).
\]
Substitute \(a_\ell(x) = \sum_i \phi_i f_i(x)\) and the expanded form of \(\hat y^{(B)}(x)\):
\begin{align*}
\nabla_{w^{(B)}}\mathcal L_B(x)
&= \Bigg(\sum_j (w^{(B)\top}\phi_j)\, f_j(x) - y^{(B)}(x)\Bigg) \sum_i \phi_i f_i(x).
\end{align*}
Taking the expectation over \(x\sim\mathcal D_B\) and distributing the summation over \(i\):
\begin{align*}
\mathbb{E}_{\mathcal D_B}[\nabla_{w^{(B)}}\mathcal L_B]
&= \sum_i \phi_i \Bigg( \sum_j (w^{(B)\top}\phi_j)\, \mathbb{E}[f_j(x)f_i(x)] - \mathbb{E}[y^{(B)}(x) f_i(x)] \Bigg).
\end{align*}
Applying the definitions of \(\gamma_j^{(B)}\), \(\Sigma_{ij}^{(B)}\), and \(\beta_i^{(B)}\) yields:
\[
\mathbb{E}_{\mathcal D_B}[\nabla_{w^{(B)}}\mathcal L_B]
= \sum_i \Bigg(\sum_j \gamma_j^{(B)}\Sigma_{ij}^{(B)} - \beta_i^{(B)}\Bigg)\, \phi_i.
\]
Finally, a gradient-descent step with step size \(\eta\) updates \(\phi_i\) and \(w^{(B)}\) by
\begin{align*}
\Delta\phi_i &= -\eta\,\mathbb{E}_{\mathcal D_B}\big[\nabla_{\phi_i}\mathcal L_B\big]
= -\eta\Bigg(\sum_j \gamma_j^{(B)}\Sigma_{ij}^{(B)} - \beta_i^{(B)}\Bigg)\, w^{(B)}, \\
\Delta w^{(B)} &= -\eta\,\mathbb{E}_{\mathcal D_B}\big[\nabla_{w^{(B)}}\mathcal L_B\big] = -\eta \sum_{i} \left( \sum_{j}\gamma_{j}^{(B)}\Sigma_{ij}^{(B)} - \beta_{i}^{(B)} \right) \phi_{i},
\end{align*}
which is \eqref{eq:updates}.
\end{proof}
\section{Extended analysis of feature updates}

\subsection{Feature updates with shared probes}\label{apx:feature_updates_shared}
The main analysis focuses on the case where a different probe is used at each task. Here, we extend the analysis to the case in which all probes are shared across tasks. When focusing on the penultimate layer, this would correspond to a \textbf{class-incremental learning (CIL)} setting with a shared classifier over all classes.

We assume that the two adjacent tasks $A$ and $B$ contain samples from disjoint class sets. For each class $c$, samples belonging to that class have a nonzero target $y_c(\bm x) \ne 0$, while samples from other classes have target $y_c(\bm x)=0$. Importantly, old classes are not masked during training on task $B$; instead, they receive zero targets.

Under this setup, gradient descent on task $B$ induces an additional data-dependent term on the feature vectors, arising from classes in previous tasks. This term depends on feature correlations under $\mathcal{D}_B$ and on the alignment between features and old-class probes, and it reduces the contribution of features to old-class predictions. In other words, \textbf{this additional term acts as a suppressive pressure on features predictive of Task $A$ classes and activated under $\mathcal{D}_B$}. Formally:

\begin{proposition}[Expected feature vector update during Task \(B\) training with shared probes]\label{lemma:phi_update_cil_B}
Let \(\mathcal{C}_A\) and \(\mathcal{C}_B\) be disjoint sets of classes, where \(\mathcal{C}_A\) was learned previously in Task A. Let the shared probe matrix be \(W = [w_1, w_2, \dots, w_K]\). Under Task \(B\) training (\(\bm x \sim \mathcal{D}_B\)), no classes from Task $A$ appear and thus the targets for Task \(A\) classes are strictly zero: \(y_c^{(B)}(\bm x) = 0\) for all \(c \in \mathcal{C}_A\). 

Define the feature correlation as \(\Sigma_{i,j}^{(B)} := \mathbb{E}_{\bm x\sim\mathcal D_B}[f_i(\bm x)f_j(\bm x)]\), the probe-feature alignment as \(\gamma_{j,c} := w_c^\top \bm \phi_j\), and the feature contribution as \(\beta_{i,c}^{(B)} := \mathbb{E}_{\bm x\sim\mathcal D_B}[y_c^{(B)}(\bm x) f_i(\bm x)]\).

The expected gradient-descent update on feature vector \(\bm \phi_i\) during Task \(B\) is:
\begin{equation}\label{eq:phi_update_lemma_cil_B}
\mathbb{E}_{\bm x\sim\mathcal D_B}[\Delta\bm \phi_i] = -\eta \left[ 
\underbrace{\sum_{c \in \mathcal{C}_B} \Bigg(\sum_j \gamma_{j,c}\Sigma_{ij}^{(B)} - \beta_{i,c}^{(B)}\Bigg) w_c}_{\text{Task B Learning}} 
+ \underbrace{\sum_{c \in \mathcal{C}_A} \Bigg(\sum_j \gamma_{j,c}\Sigma_{ij}^{(B)}\Bigg) w_c}_{\text{Task A Suppression}}
\right].
\end{equation}
\end{proposition}

\begin{proof}
We define the per-example squared loss across all \(K\) classes for Task \(B\) as:
\[
\mathcal{L}_B(\bm x) = \frac{1}{2} \sum_{c=1}^K \big(\hat{y}_c(\bm x) - y_c^{(B)}(\bm x)\big)^2, \qquad \hat{y}_c(\bm x) = w_c^\top a_\ell(\bm x).
\]
By the chain rule, the gradient with respect to \(\bm \phi_i\) is obtained by summing the contribution from each class output. Since \(a_\ell(\bm x) = \sum_j \bm \phi_j f_j(\bm x)\), we have \(\nabla_{\bm \phi_i} \hat{y}_c(\bm x) = w_c f_i(\bm x)\). Therefore:
\[
\nabla_{\bm \phi_i} \mathcal{L}_B(\bm x) = \sum_{c=1}^K \big(\hat{y}_c(\bm x) - y_c^{(B)}(\bm x)\big) w_c f_i(\bm x).
\]
Expanding \(\hat{y}_c(\bm x) = \sum_j (w_c^\top \bm \phi_j) f_j(\bm x) = \sum_j \gamma_{j,c} f_j(\bm x)\) and distributing \(f_i(\bm x) w_c\):
\begin{align*}
\nabla_{\bm \phi_i} \mathcal{L}_B(\bm x) &= \sum_{c=1}^K \Bigg( \sum_j \gamma_{j,c} f_j(\bm x) - y_c^{(B)}(\bm x) \Bigg) w_c f_i(\bm x) \\
&= \sum_{c=1}^K \Bigg( \sum_j \gamma_{j,c} f_j(\bm x) f_i(\bm x) - y_c^{(B)}(\bm x) f_i(\bm x) \Bigg) w_c.
\end{align*}
Taking the expectation over \(\bm x \sim \mathcal{D}_B\) and applying linearity of expectation:
\[
\mathbb{E}_{\bm x\sim\mathcal{D}_B}[\nabla_{\bm \phi_i} \mathcal{L}_B(\bm x)] = \sum_{c=1}^K \Bigg( \sum_j \gamma_{j,c} \Sigma_{ij}^{(B)} - \beta_{i,c}^{(B)} \Bigg) w_c,
\]
where we have used the definitions \(\Sigma_{ij}^{(B)} = \mathbb{E}_{\bm x\sim\mathcal{D}_B}[f_i(\bm x)f_j(\bm x)]\) and \(\beta_{i,c}^{(B)} = \mathbb{E}_{\bm x\sim\mathcal{D}_B}[y_c^{(B)}(\bm x) f_i(\bm x)]\).

Since we are training on Task B, the labels for previously learned Task A classes are zero: \(y_c^{(B)}(\bm x) = 0\) for all \(c \in \mathcal{C}_A\). Consequently, \(\beta_{i,c}^{(B)} = \mathbb{E}_{\bm x\sim\mathcal{D}_B}[0 \cdot f_i(\bm x)] = 0\) for \(c \in \mathcal{C}_A\).

Partitioning the sum over all \(K\) classes into \(\mathcal{C}_B\) and \(\mathcal{C}_A\):
\begin{align*}
\mathbb{E}_{\bm x\sim\mathcal{D}_B}[\nabla_{\bm \phi_i} \mathcal{L}_B(\bm x)] &= \sum_{c \in \mathcal{C}_B} \Bigg( \sum_j \gamma_{j,c} \Sigma_{ij}^{(B)} - \beta_{i,c}^{(B)} \Bigg) w_c \\
&\quad + \sum_{c \in \mathcal{C}_A} \Bigg( \sum_j \gamma_{j,c} \Sigma_{ij}^{(B)} - 0 \Bigg) w_c.
\end{align*}
The gradient-descent update with learning rate \(\eta\) is
\[
\mathbb{E}_{\bm x\sim\mathcal D_B}[\Delta\bm \phi_i] = -\eta\,\mathbb{E}_{\bm x\sim\mathcal D_B}\big[\nabla_{\bm \phi_i}\mathcal L_B(\bm x)\big],
\]
which yields \cref{eq:phi_update_lemma_cil_B}.
\end{proof}

\Cref{eq:phi_update_lemma_cil_B} isolates the mechanism of feature forgetting and readout misalignment with shared probes: the second summation acts as a suppressive term, actively pushing features away from Task A probes to satisfy the \(y_c^{(B)}=0\) targets. 
% In a fixed-probe scenario (as in the main analysis), the feature matrix $\Phi$ must absorb the entirety of this suppression load, placing the full burden of this suppressive pressure on the feature representations. However, as discussed in \cref{apx:probe_coadapt}, under simultaneous probe co-adaptation, the Task A weight vectors $w_c$ (for \(c \in \mathcal{C}_A\)) also update, naturally decaying or rotating away from the active Task B features. By sharing this gradient load, the degradation of the old classifier weights reduces the impact of optimization on underlying features. Thus, while fixed-probe analysis is useful for isolating feature dynamics, it represents a worst-case scenario that overestimates the magnitude of representation degradation.

\subsection{Feature updates under cross-entropy loss}\label{apx:feature_updates_shared_ce}

We now extend the analysis to the standard classification setting using the cross-entropy (CE) loss with a softmax activation, to better understand how a different loss function affects the results in the penultimate layer of a model.

We define the per-example CE loss as \(\mathcal{L}_{CE}(\bm x) = - \sum_{c=1}^K y_c^{(B)}(\bm x) \log p_c(\bm x)\), where the class probability \(p_c(\bm x)\) is given by the softmax of the logits:
\begin{equation}
    p_c(\bm x) = \frac{\exp\left(w_c^\top a_\ell(\bm x)\right)}{\sum_{k=1}^K \exp\left(w_k^\top a_\ell(\bm x)\right)}.
\end{equation}
Recall that for Task $B$ samples, targets for Task $A$ classes are zero: \(y_c^{(B)}(\bm x)=0\) for all \(c \in \mathcal{C}_A\). Taking the expectation over $\mathcal{D}_B$, the update rule becomes:

\begin{proposition}[Feature vector update under cross-entropy]\label{lemma:phi_update_ce}
Let the loss function be the cross-entropy loss \(\mathcal{L}_{CE}(\bm x) = - \sum_{k=1}^K y_k^{(B)}(\bm x) \log p_k(\bm x)\), where \(p_k(\bm x)\) is the softmax output. Under the same task setup as \cref{lemma:phi_update_cil_B} (disjoint class sets with zero targets for old classes \(c \in \mathcal{C}_A\)), the expected update for feature vector $\bm \phi_i$ is:
\begin{equation}\label{eq:phi_update_ce}
    \mathbb{E}_{\bm x\sim\mathcal D_B}[\Delta\bm \phi_i] = -\eta \left[ \sum_{c \in \mathcal{C}_B} \Bigg(\mathbb{E}_{\bm x\sim\mathcal D_B}[p_c(\bm x) f_i(\bm x)] - \beta_{i,c}^{(B)}\Bigg) w_c + \sum_{c \in \mathcal{C}_A} \Bigg(\mathbb{E}_{\bm x\sim\mathcal D_B}[p_c(\bm x) f_i(\bm x)]\Bigg) w_c \right].
\end{equation}
\end{proposition}

\begin{proof}
The gradient of the cross-entropy loss with respect to the logit $\hat{y}_c(\bm x)$ is the standard softmax error signal:
\[
\frac{\partial \mathcal{L}_{CE}(\bm x)}{\partial \hat{y}_c(\bm x)} = p_c(\bm x) - y_c^{(B)}(\bm x).
\]
Recall that the logit decomposes as \(\hat{y}_c(\bm x) = w_c^\top a_\ell(\bm x) = w_c^\top \sum_j \bm \phi_j f_j(\bm x)\). The gradient of \(\hat{y}_c(\bm x)\) with respect to the feature vector $\bm \phi_i$ is:
\[
\nabla_{\bm \phi_i} \hat{y}_c(\bm x) = w_c f_i(\bm x).
\]
Applying the chain rule and summing over all $K$ classes:
\begin{align*}
    \nabla_{\bm \phi_i} \mathcal{L}_{CE}(\bm x) &= \sum_{c=1}^K \frac{\partial \mathcal{L}_{CE}(\bm x)}{\partial \hat{y}_c(\bm x)} \cdot \nabla_{\bm \phi_i} \hat{y}_c(\bm x) \\
    &= \sum_{c=1}^K \big(p_c(\bm x) - y_c^{(B)}(\bm x)\big) w_c f_i(\bm x).
\end{align*}
Taking the expectation over the data distribution $\mathcal{D}_B$ and partitioning the sum into current task classes $\mathcal{C}_B$ and old task classes $\mathcal{C}_A$:

For current task classes (\(c \in \mathcal{C}_B\)):
\[
\mathbb{E}_{\bm x\sim\mathcal{D}_B}\Big[\big(p_c(\bm x) - y_c^{(B)}(\bm x)\big) f_i(\bm x)\Big] w_c = \Big(\mathbb{E}_{\bm x\sim\mathcal{D}_B}[p_c(\bm x) f_i(\bm x)] - \beta_{i,c}^{(B)}\Big) w_c,
\]
where we used the definition \(\beta_{i,c}^{(B)} = \mathbb{E}_{\bm x\sim\mathcal{D}_B}[y_c^{(B)}(\bm x) f_i(\bm x)]\).

For old task classes (\(c \in \mathcal{C}_A\)), the targets are zero: \(y_c^{(B)}(\bm x) = 0\). Thus:
\[
\mathbb{E}_{\bm x\sim\mathcal{D}_B}\Big[\big(p_c(\bm x) - 0\big) f_i(\bm x)\Big] w_c = \mathbb{E}_{\bm x\sim\mathcal{D}_B}[p_c(\bm x) f_i(\bm x)] w_c.
\]
The gradient descent update \(\mathbb{E}_{\bm x\sim\mathcal{D}_B}[\Delta \bm \phi_i] = -\eta \mathbb{E}_{\bm x\sim\mathcal{D}_B}[\nabla_{\bm \phi_i} \mathcal{L}_{CE}(\bm x)]\) yields \cref{eq:phi_update_ce}.
\end{proof}

\Cref{eq:phi_update_ce} preserves the structural form of the MSE update in \cref{lemma:phi_update_cil_B}. The first summation drives alignment with the new task (incorporating the feature contribution $\beta_{i,c}^{(B)}$), while the second summation acts as the suppressive pressure from old tasks. Note, however, that under CE, feature correlations are implicit via $p_c(\bm x)$, and suppression depends on the full logit distribution through the softmax.

The nature of the suppression, however, differs fundamentally. In the linear MSE case, the suppressive term depends on the raw logits (\(\hat{y}_c(\bm x)\)), scaling linearly with feature activations through \(\sum_j \gamma_{j,c}\Sigma_{ij}^{(B)}\). Under CE, the suppressive term for an old class \(c \in \mathcal{C}_A\) is weighted by the predicted probability \(p_c(\bm x)\). Consequently, suppressive pressure in CE is dynamic and probability-modulated: it is strong only when a Task $B$ sample is "confused" for an old class (i.e., when \(p_c(\bm x)\) is non-negligible). If the new task features are approximately orthogonal to the old class probes, or if the model is confident in its Task $B$ predictions, then \(p_c(\bm x) \to 0\) for \(c \in \mathcal{C}_A\), and the suppressive gradient becomes negligible. This suggests that \textbf{CE loss may offer a milder form of feature forgetting compared to MSE}, provided the classes are sufficiently separable in the representation space and the model maintains high confidence on new task predictions.
\section{Experimental setup of the feature-reader model}\label{apx:experimental_setup_featurereader}

\subsection{Feature-reader Model Details}

We implemented the feature-reader model in PyTorch~\citep{imambi2021pytorch}. Unless stated otherwise, the model consists of a shared 1-layer encoder, $\Phi$, which maps 80 features to 20 activation dimensions, and five probes ($W_t$) per task.
\begin{itemize}
    \item \textbf{Task definition}: Each task is defined by a unique contribution vector, $\beta_t$, which specifies a sparse subset of features relevant to that task. $\beta_t$ is randomly sampled from a standard normal distribution.
    \item \textbf{Data generation}: We generated $100,000$ synthetic data samples per task by sampling feature activations uniformly in the range [0, 1] with 90\% sparsity (feature $i$ is 0 with a probability of 90\%). In the `none' scenario, this was followed by masking out features from other tasks.
    \item \textbf{Training}: The model was trained sequentially on five tasks using batch gradient descent with the Adam optimizer and $lr=0.001$. We minimized the Mean Squared Error (MSE) loss. We trained the model for $10,000$ epochs at each task to ensure convergence. We clipped gradient norms to 1 to avoid instability issues. All results reported are the average of 5 independent seeds.
\end{itemize}

\subsection{Forgetting Metric (F) Formulation}

To quantify forgetting, we measure the degradation of a given metric (e.g., accuracy, normalized capacity) for all previously learned tasks. 

For any task $i$ and any subsequent task $t$ (where $t > i$), we first compute the relative performance ratio, $R_{i,t}$:
\begin{equation}
    R_{i,t} = \frac{M_{i,t}}{M_{i,i}},
\end{equation}
where$M_{i,i}$ is the value of the metric for task $i$ immediately after it was trained and $M_{i,t}$ is the value of the same metric for task $i$ after the model has finished training on task $t$.

The forgetting value for task $i$ is $1 - R_{i,t}$. A value of 1 indicates complete forgetting, meaning the metric's value has dropped to zero relative to its original level. 

The final \textbf{forgetting (F)} score reported in our plots at step $t$ is the average forgetting across all previously learned tasks $i < t$:
\begin{equation}
    F = \frac{1}{t-1} \sum_{i=1}^{t-1} (1 - R_{i,t}).
\end{equation}
Before this calculation, all metrics are first averaged over the features associated with a specific task.

% We implemented the feature-reader model in PyTorch to empirically validate the analytical results. The model consists of a shared encoder $\Phi \in \mathbb{R}^{m \times n}$ with configurable depth and nonlinearity, and task-specific linear heads $w_t$. Each task is defined by a subset of active features and a contribution vector $\beta_t$, determining how strongly each feature predicts the label. Synthetic data were generated by sampling feature activations uniformly in $[0,1]$, applying sample-level sparsity, and masking inactive features.

% Tasks were trained sequentially with Adam and MSE loss, updating both encoder and head parameters. After each task, we stored $\Phi^{(t)}$ to track representational drift. We measured per-task \emph{capacity}, \emph{norms}, \emph{probe sensitivity} ($\gamma_i$), and \emph{accuracy} ($1/(1+\text{MSE})$) across snapshots.

\section{Extended analysis of the feature-reader experiments}\label{apx:extended_toy_experiments}

\subsection{Task similarity under the lens of our mechanistic framework}

\begin{figure}[htb]
    \centering
    \vspace{-6pt}
    \begin{subfigure}{1\columnwidth}
        \hspace{-4pt}
        \begin{subfigure}{0.249\columnwidth}
        \includegraphics{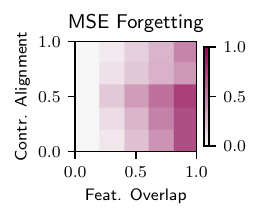}
        \end{subfigure}%
        \hspace{4pt}
        \begin{subfigure}{0.249\columnwidth}
        \includegraphics{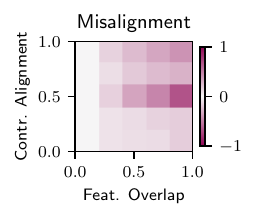}
        \end{subfigure}%
        \begin{subfigure}{0.249\columnwidth}
        \includegraphics{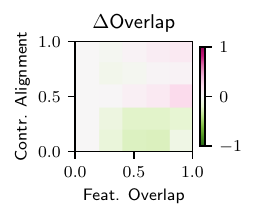}
        \end{subfigure}%
        \begin{subfigure}{0.249\columnwidth}
        \includegraphics{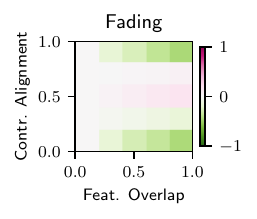}
        \end{subfigure}%
    \end{subfigure}
    
    \begin{subfigure}{1\columnwidth}
        \hspace{-4pt}
        \begin{subfigure}{0.249\columnwidth}
        \includegraphics{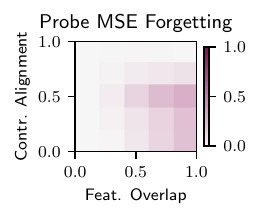}
        \end{subfigure}%
        \hspace{4pt}
        \begin{subfigure}{0.249\columnwidth}
        \includegraphics{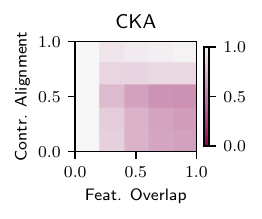}
        \end{subfigure}%
        \hspace{4pt}
        \begin{subfigure}{0.249\columnwidth}
        \includegraphics{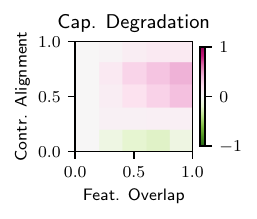}
        \end{subfigure}%
    \end{subfigure}

    \vspace{-10pt}
    \caption{\textbf{Forgetting, capacity degradation, and CKA metrics for different similarities.} 
    }
    \label{fig:task_similarity_heatmaps}
\end{figure}

We systematically vary two dimensions of task similarity—feature overlap and contribution alignment—and measure the resulting forgetting and capacity dynamics. Feature overlap (Feat. Overlap) controls the degree to which the same features are co-activated across tasks, interpolating between the disjoint (\cref{cor:best_case}) and fully shared (\cref{cor:worst_case}) regimes. Contribution alignment (Contr. Alignment) controls whether shared features carry the same predictive role across tasks, ranging from fully anti-aligned (0.0) to fully aligned (1.0) contributions.

\paragraph{Accuracy forgetting follows a U-shaped pattern in contribution alignment} At zero feature overlap (leftmost column), accuracy forgetting is uniformly near zero regardless of contribution alignment---consistent with \cref{cor:best_case}, which guarantees that task-A features receive no gradient pressure when $\Sigma^{(B)}_{ij}=0$. As overlap increases, forgetting grows substantially, but the relationship with contribution alignment is non-monotonic. At full overlap (feature overlap = 1.0), uninformative contributions $\beta_i^{(B)} \approx 0$ (contribution alignment $\approx 0.5$) yield the worst forgetting, reaching 0.36, compared to 0.33 at zero alignment and only 0.21 at perfect alignment. This U-shaped profile aligns with our discussion in \cref{sec:dynamics}: anti-aligned contributions cause high forgetting. However, in practice, uninformative features whose contribution is $\beta_i^{(B)} \approx 0$ do not have any incentive to be kept "alive" and seem to suffer from the worst gradient pressure.

\paragraph{Contribution alignment preserves feature structure} When tasks share features and assign them identical contributions (alignment = 1.0), the model essentially reinforces the same features across tasks. The data supports this: at (overlap=1.0, alignment=1.0), while standard accuracy forgetting is 0.21, the probe accuracy forgetting is near zero. This reveals that the underlying feature geometry is almost entirely preserved. The residual drop in standard accuracy is not due to a capacity degradation, but rather readout scale misalignment driven by negative fading. Specifically, the feature norms significantly increase as they are reinforced by the new task. The original task probe, calibrated to smaller feature norms, now misreads these amplified features. Thus, the features remain perfectly intact and readable, but their altered scale induces standard performance loss. 

\paragraph{Probe accuracy and capacity metrics jointly identify feature geometry rearrangement} Probe accuracy decays substantially less than standard accuracy in most regimes---for instance, at (overlap=0.5, alignment=0.5), standard accuracy forgetting is 0.26 while probe accuracy forgetting is only 0.12---indicating that much of what looks like forgetting is simply readout misalignment rather than true information loss. However, at extreme conflicting alignments (overlap=1.0, alignment=0.0), a particular failure mode emerges: probe accuracy still drops (0.16) even though feature capacity and norms \textit{increase} (norm forgetting = $-$0.436). When tasks directly conflict, strong gradients actively force a geometric rearrangement of the shared representations to satisfy the new task's optimal packing constraints. Note that, in this experiment, superposition is forced when more than 20 features are active (the encoder has 20 neurons). The features themselves remain highly utilized, driving the increased capacity, but their relative entanglement shifts. This systemic rearrangement destroys the specific interference-canceling geometry that the original probe relied upon, making it more difficult to isolate the target feature from newly entangled noise. For example, imagine a task where the probe was positively aligned with one feature and antialigned with a second feature. If, after a new task, the dot product of these features becomes positive, the old task structure cannot be recovered. Therefore, the original knowledge is not necessarily erased, but actively repacked into a geometry that compromises the old readout.

\paragraph{Our mechanistic metrics complement CKA and linear probing} While CKA is highly effective at tracking global structural rotations, it can understate performance loss driven by scaling effects, highlighting the value of a complementary mechanistic vocabulary. For instance, at (overlap=1.0, alignment=1.0), CKA remains remarkably high, correctly indicating that the feature directions are structurally stable. However, standard accuracy forgetting is still 0.21 in this regime---a discrepancy that CKA alone cannot explain. Our framework explicitly tracks changes in feature norms, and we reveal that the performance drop is driven by significant feature amplification (norm forgetting = -0.435), causing readout scale misalignment. By decomposing representation changes into primitives, capacity degradation (overlap and fading), and misalignment, we provide a richer diagnostic lens for scenarios where the global geometry appears stable but functional performance degrades.

\section{Introduction to crosscoders}\label{apx:introduction_crosscoders}
% Intro SAEs
%~\citep{OLSHAUSEN19973311}
\subsection{Sparse Autoencoders}

The linear representation hypothesis~(\Cref{eq:activation_linear}) posits that features are encoded as directions in activation space, while the superposition hypothesis suggests the number of underlying features exceeds the neuron count ($d_{\text{model}}$), making features difficult to disentangle.

\emph{Sparse Autoencoders (SAEs)} are employed to \emph{decouple} these superposed features and uncover the individual features learned by a neural network. An SAE maps a layer representation $a_\ell(x) \in \mathbb{R}^{d_{\text{model}}}$ to a \emph{higher-dimensional, sparse latent space} ($d_{\text{cross}} > d_{\text{model}}$). In this latent space, each dimension is encouraged to be \emph{monosemantic}, corresponding to a single feature.

The SAE operates via an encoding-decoding process:

\begin{align}\label{eq:standard_sae_apx}
    f_{\text{enc}}(x) &= g(\mW_{\text{enc}}a_\ell(x) + b_{\text{enc}})\\
    \text{SAE}(x) &= \mW_{\text{dec}} f_{\text{enc}}(x) + b_{\text{dec}}
\end{align}

where $\mW_{\text{enc}} \in \mathbb{R}^{d_{\text{cross}} \times d_{\text{model}}}$ and $\mW_{\text{dec}} \in \mathbb{R}^{d_{\text{model}} \times d_{\text{cross}}}$ are the learnable encoder and decoder matrices (the \emph{dictionary}), respectively. The encoder produces feature activations $f_{\text{enc}}(x) \in \mathbb{R}^{d_{\text{cross}}}$, using ReLU and $\text{TopK}$ activation functions $g(\cdot)$~\citep{gao2025scaling}.

SAEs are trained to accurately reconstruct the original representation $a_\ell(x)$ while enforcing $\ell_1$ \emph{sparsity} on the feature activations. The loss function defines this:

\begin{equation}\label{eq:loss_sae}
    \mathcal{L}(x) = \norm{a_\ell(x) - \text{SAE}(x)}_2 + \alpha\norm{f_{\text{enc}}(x)}_1.
\end{equation}

The sparsity constraint ensures that, once trained, $a_\ell(x)$ is approximated as a \emph{sparse linear combination} of the columns of the learned dictionary $\mW_{\text{dec}}$, from which interpretable and monosemantic features are often extracted~\citep{bricken2023monosemanticity,huben2024sparse}.

\paragraph{SAEs and our conceptual framework}

We use SAEs to extend the feature analysis to larger neural networks. The components of the trained SAE directly map to the concepts in our framework:
\begin{itemize}
    \item The $i$-th component of the feature activation, $f_{\text{enc},i}(x)$, is interpreted as the feature activation $f_i(x)$.
    \item The $i$-th column of the decoder matrix, $\mW_{\text{dec},i}$, is analogous to the feature vector $\phi_i$, as it defines feature $i$'s contribution to the reconstructed activation.
\end{itemize}

\subsection{Crosscoders}
\emph{Crosscoders}~\citep{lindsey2025crosscoders} are a specialized variant of SAEs designed to study features encoded across multiple layers and for \emph{model diffing}, which is the study of how different models encode features. This approach is particularly valuable for analyzing sequential changes, such as in a \emph{continual learning setup} with a set of models $\mathcal{M} = \{m_1, m_2,...,m_T\}$, corresponding to models trained on $T$ sequential tasks.
The core distinction of a crosscoder is the \emph{shared encoding step}. Representations from different models $a_{\ell}^{m_t}(x)$ (at layer $\ell$ for model $m_t$) for the same input $x$ are aggregated and mapped into a \emph{single, shared latent feature space}:

\begin{equation}
f_{\text{enc}}(x) = g\Bigl( \sum_{{m_t} \in \mathcal{M}} \mW_{\text{enc}}^{m_t}
a_{\ell}^{m_t}(x) + b_{\text{enc}}\Bigr),
\end{equation}

with $\mW_{\text{enc}}^{m_t}$ and $b_{\text{enc}}$ denoting the encoder weights for model $m_t$, and $a_{\ell}^{m_t}$ the representation of model $m_t$ at layer $l$. The decoding step uses individual decoders to reconstruct each model's original representation from the shared activations:

\begin{equation}
    \text{CROSS}^{m_t}(a^{m_t}_\ell(x)) = \mW^{m_t}_{\text{dec}}f_{\text{enc}}(x) + b_{\text{dec}}^{m_t},
\end{equation}

where $b_{\text{dec}}^{m_t}$ is the corresponding model-specific bias. A schematic overview of the process is provided in~\Cref{fig:crosscoder_diagram}. The training objective is a combination of reconstruction loss and a regularization term:

\begin{equation}\label{eq:crosscoder}
\mathcal{L}(x) = \sum_{{m_t} \in \mathcal{M}} \norm{a^{m_t}_\ell(x) - \text{CROSS}^{m_t}(a^{m_t}_\ell(x))}_2 
    + \lambda \sum_i f_{\text{enc}_i}(x) \sum_{{m_t} \in \mathcal{M}} \|W^{m_t}_{\text{dec}, i}\|.
\end{equation}

The second term, $\lambda \sum_i f_{\text{enc}_i}(x) \sum_{{m_t} \in \mathcal{M}} \|\mW^{m_t}_{\text{dec}, i}\|_2$, encourages feature compression. It pushes the decoder weights for feature $i$ ($\mW^{m_t}_{\text{dec}, i}$) toward zero norm if feature $i$ is inactive or non-shared across a subset of models. This regularization helps \emph{isolate features} that are unique to certain models or tasks.

\paragraph{Application to continual learning}
In the context of this work, the crosscoder analysis enables us to study the \textbf{evolution} of the decoder vector $\mW^{m_t}_{\text{dec}, i}$ across sequential tasks $m_t$. Similar to the analysis of the feature vector $\phi_i$ in the conceptual framework~(\Cref{sec:conceptualization}), we can track the transformation of features and the allocation of feature capacity across the continual learning process.

\section{Setup for the ViT and crosscoder experiment}\label{apx:crosscoders_experimental_details}

This section outlines the experimental setup for the continual learning experiments on the CIFAR-10 dataset, including the Vision Transformer (ViT) and Top-K Sparse Autoencoder (TopKSAE) training procedures.

\subsection{Dataset and Task Protocol}

We use the CIFAR-10 dataset, which consists of 60,000 32x32 color images in 10 classes, partitioned into a 50,000-image training set and a 10,000-image test set. For our experiments, all images are resized to 224x224.

For our continual learning benchmark, we use the \textbf{Split CIFAR-10} protocol. The 10 classes of CIFAR-10 are split into 5 sequential tasks, with each task containing 2 distinct classes. The model is trained sequentially on these tasks. The specific class sequence used for the experiments is:
\begin{itemize}
    \item \textbf{Task 1:} Classes \{5, 9\} (dog, truck)
    \item \textbf{Task 2:} Classes \{3, 1\} (cat, automobile)
    \item \textbf{Task 3:} Classes \{7, 2\} (horse, bird)
    \item \textbf{Task 4:} Classes \{6, 4\} (frog, deer)
    \item \textbf{Task 5:} Classes \{0, 8\} (airplane, ship)
\end{itemize}

\subsection{Vision Transformer (ViT) Training}

\subsubsection{Model Architecture}
Our base model is a Vision Transformer (ViT) with the following architecture:
\begin{itemize}
    \item \textbf{Image Size:} 224x224
    \item \textbf{Patch Size:} 16x16
    \item \textbf{Embedding Dimension (\texttt{hidden\_size}):} 128
    \item \textbf{Number of Layers (Depth):} 6
    \item \textbf{Number of Attention Heads:} 4
    \item \textbf{MLP Dimension (\texttt{intermediate\_size}):} 256
    \item \textbf{Output Head:} A separate linear layer is trained for each task, mapping the final [CLS] token representation to the 2 classes of that task.
\end{itemize}

\subsubsection{Training Setup}
The model is trained on each of the 5 tasks sequentially using a standard fine-tuning approach. For each new task, the model weights are initialized from the checkpoint of the previous task, and a new classification head is trained.
\begin{itemize}
    \item \textbf{Optimizer:} Adam
    \item \textbf{Learning Rate:} 1e-4
    \item \textbf{Weight Decay:} 0.0
    \item \textbf{Batch Size:} 128
    \item \textbf{Epochs per Task:} 100
    \item \textbf{Learning Rate Scheduler:} Cosine Annealing
\end{itemize}

\subsection{Top-K Sparse Autoencoder (TopKSAE) Setup}

We train and apply Top-K Sparse Autoencoders (which we refer to as "crosscoders") to the token activations of the ViT.

\subsubsection{Training}
The TopK SAEs are trained only once after the completion of the last task. They are trained on activations extracted from the ViT using the full training data from all tasks.
\begin{itemize}
    \item \textbf{Target Layers:} We train separate SAEs for the outputs of the final, third-to-last, and sixth-to-last transformer blocks (corresponding to layers \texttt{-1}, \texttt{-3}, and \texttt{-6}).
    \item \textbf{SAE Architecture:}
    \begin{itemize}
        \item \textbf{Dictionary Size (\texttt{nb\_concepts}):} 192 (calculated as \texttt{d\_model} * 1.5)
        \item \textbf{Sparsity:} We use a Top-K activation function with a \textbf{sparse factor of 6}, retaining only the six largest feature activations and setting all others to zero.
    \end{itemize}
    \item \textbf{SAE Training Hyperparameters:}
    \begin{itemize}
        \item \textbf{Optimizer:} Adam
        \item \textbf{Learning Rate:} 5e-4
        \item \textbf{Batch Size:} 256
        \item \textbf{Epochs:} 3
        \item \textbf{L1 Regularization:} max coefficient of 0.001, linearly warmed up over the first 5\% of training
    \end{itemize}
\end{itemize}

\subsection{Reproducibility}

All experiments were conducted across \textbf{4 different random seeds} to ensure the robustness of our findings.
\section{An example of a case study with a ViT (class-incremental)}\label{apx:class-incremental}

\begin{figure}[t]
    \centering    
    \vspace{-6pt}
    
    % Five plots in a single row
    \begin{subfigure}{0.19\columnwidth}
        \centering
        \includegraphics[width=\linewidth]{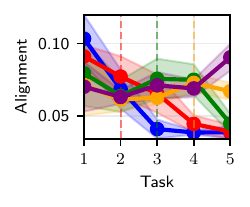}
    \end{subfigure}
    \hfill
    \begin{subfigure}{0.19\columnwidth}
        \centering
        \includegraphics[width=\linewidth]{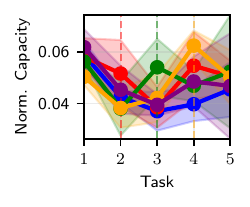}
    \end{subfigure}
    \hfill
    \begin{subfigure}{0.19\columnwidth}
        \centering
        \includegraphics[width=\linewidth]{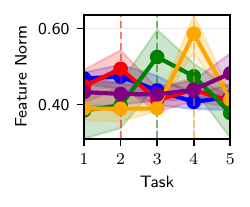}
    \end{subfigure}
    \hfill
    \begin{subfigure}{0.19\columnwidth}
        \centering
        \includegraphics[width=\linewidth]{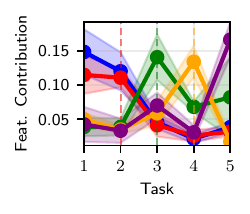}
    \end{subfigure}
    \hfill
    \begin{subfigure}{0.19\columnwidth}
        \centering
        \includegraphics[width=\linewidth]{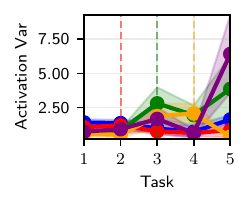}
    \end{subfigure}

    \vspace{-8pt}

    % Legend centered below the row
    \begin{subfigure}{1\columnwidth}
        \centering
        \makebox[\textwidth][c]{
            % \hspace{-1.5\columnwidth}
            \begin{minipage}{0.42\columnwidth}
                \flushright
                \small Metrics for task:
            \end{minipage}%
            \begin{minipage}{0.58\columnwidth}
                \includegraphics{assets/crosscoders/results/seq_layer_m1/legend.pdf}
            \end{minipage}%
        }
    \end{subfigure}
    \vspace{-10pt}
    
    \caption{\textbf{Metrics evolution across tasks for the penultimate layer under class-incremental training.} Shown is the average value for each metric over the top-5 most important features of each task.}
    \label{fig:results_ci_crosscoder}
\end{figure}

We also repeated the case study experiment with the same configuration (\cref{apx:crosscoders_experimental_details}) in a class-incremental setup, where we did not use any tricks, such as masking for previous classes. Interestingly, \cref{fig:results_ci_crosscoder} shows that the feature-probe alignment ($\gamma$) and the norms do not drop as sharply as in the task-incremental setup experiment. On the other hand, the normalized capacities (Norm. Capacity) seem to be more sensitive to new tasks, suggesting that overlap is an issue during class-incremental learning. Compared to the task-incremental experiment, $\gamma$ tends to be suppressed more gradually, even though we found that class-incremental introduces a suppressive term in \cref{apx:feature_updates_shared}. However, the analysis of cross-entropy loss in \cref{apx:feature_updates_shared_ce} suggests that it introduces complex dynamics. We leave further analysis of cross-entropy and other losses as future work.
\section{Feature visualizations across tasks}

\begin{figure}[!hbt]
  \centering
  \includegraphics[width=0.65\textwidth]{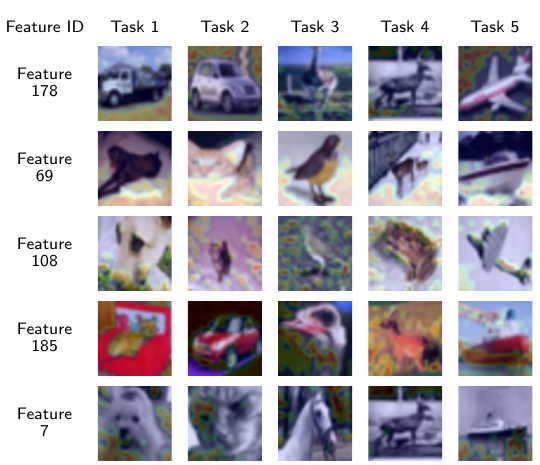}
  \caption{\textbf{Activation maps for the top-5 (descending order) most important features in task 1 across tasks for layer -1.}
  Shown are the images that most strongly activated each feature at each task. We overlay the images with their corresponding activation maps.
  }
  \label{fig:feats_m1}
\end{figure}

\begin{figure}[!hbt]
  \centering
  \includegraphics[width=0.65\textwidth]{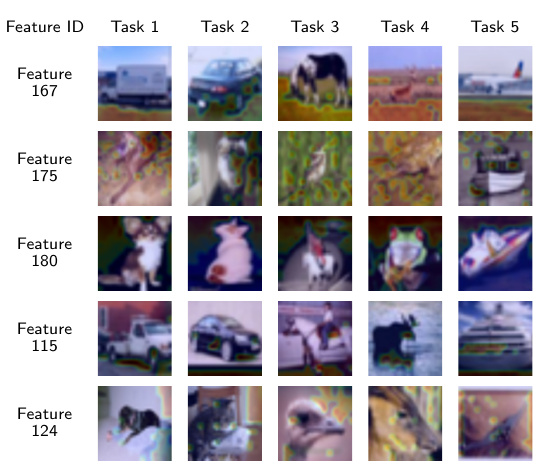}
  \caption{\textbf{Activation maps for the top-5 (descending order) most important features in task 1 across tasks for layer -3.}
  Shown are the images that most strongly activated each feature at each task. We overlay the images with their corresponding activation maps.
  }
  \label{fig:feats_m3}
\end{figure}

\begin{figure}[!hbt]
  \centering
  \includegraphics[width=0.65\textwidth]{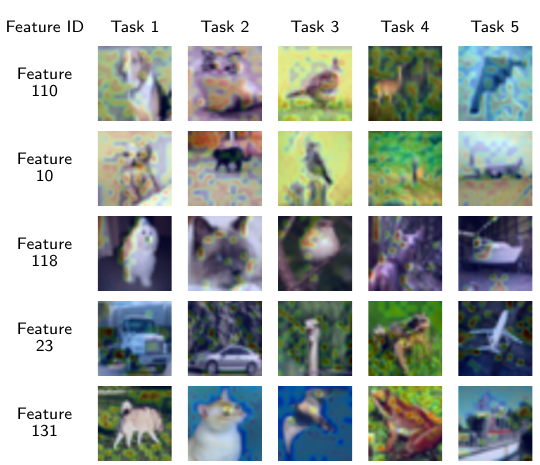}
  \caption{\textbf{Activation maps for the top-5 (descending order) most important features in task 1 across tasks for layer -6.}
  Shown are the images that most strongly activated each feature at each task. We overlay the images with their corresponding activation maps.
  }
  \label{fig:feats_m6}
\end{figure}

We visualize the activation maps of the top five important features of task 1 across all tasks for each layer. Feature IDs are layer-specific and not shared across layers. The activation maps highlight the image regions where the feature activates. Many features appear to correspond to confounders, such as asphalt (\cref{fig:feats_m1}, first row) or grass (\cref{fig:feats_m3}, first row), and these activations remain consistent across tasks, matching the worst-case scenario in \cref{cor:worst_case}. Features in shallower layers tend to be less interpretable (\cref{fig:feats_m6}). Overall, this qualitative analysis illustrates how crosscoders can be used to probe model representations, including the identification of confounding features.

\FloatBarrier

\FloatBarrier

\FloatBarrier
\section{Repeated features across layers}

\begin{figure}[hbt]
 \includegraphics[width=1\columnwidth]{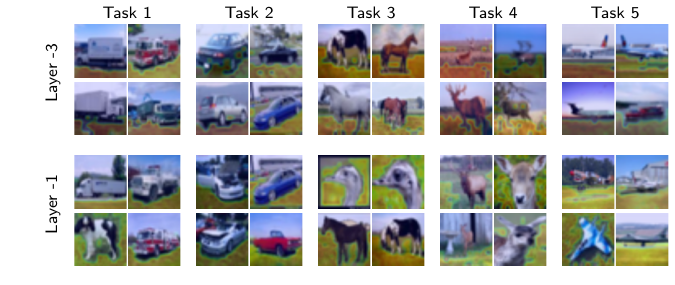}
  \caption{\textbf{Activation maps of layer -3's feature 167 learned in task 1, and layer -1's feature 131 learned in task 5.}
  Shown are the images that most strongly activated this feature at each task.
  }
  \label{fig:feat_across_layers_tasks}
\end{figure}

In~\Cref{fig:feat_across_layers_tasks} we show activation maps illustrating how two highly important features, one in the middle transformer layer (layer -3) and one in the final layer (layer -1), activate across image patches. Both features appear to represent the same underlying concept: a flat surface such as a floor or ground. This provides an example of a feature that remains constant across layers. One may think that the model has an excess of depth. However, for making a correct classification, some low-level features might need to coexist with other high-level features that do require additional processing across layers. 

This example also reveals an interesting phenomenon: a last-layer feature that is important for task 1 but whose norm has faded at task 5 (\Cref{fig:results_crosscoder}) re-emerges as an important task 5 feature in a middle layer. This might mirror the depth effects observed in \cref{sec:results_toy_depth}, where features learned earlier in the network fade as depth increases. Future continual learning methods could leverage this by restoring faded late-layer features that are also identified at earlier layers to mitigate forgetting.

\FloatBarrier

\end{document}